\pdfoutput=1
\documentclass[11pt]{article}

\usepackage{EMNLP2023}

\usepackage{times}
\usepackage{latexsym}

\usepackage[T1]{fontenc}
\usepackage[utf8]{inputenc}

\usepackage{microtype}

\usepackage{amssymb}% http://ctan.org/pkg/amssymb
\usepackage{pifont}% http://ctan.org/pkg/pifont
\newcommand{\cmark}{\ding{51}}%
\newcommand{\xmark}{\ding{55}}%

\usepackage{inconsolata}
\usepackage{float}

\usepackage{booktabs} % toprule, bottomrule

\usepackage{amsmath}
\usepackage{amsfonts}
\usepackage{graphicx}
\usepackage[most]{tcolorbox}

\newcommand{\NA}{---}
\usepackage{xcolor,colortbl}
\usepackage{paralist} % Inline lists
\definecolor{rowrow}{RGB}{164, 188, 219}
\definecolor{row}{RGB}{236,244,233}
\definecolor{rowrow}{RGB}{208,219,203}
\definecolor{transductive}{HTML}{F57251}
\definecolor{inductive}{HTML}{67AB9F}
\definecolor{polysemous}{HTML}{A4BCDB}
\definecolor{outofkg}{HTML}{C4AD9D}
\newcommand{\rpar}[1]{\vspace{1.4mm}\noindent\textbf{#1.}}

\usepackage{tikz}
\newcommand{\tikzxmark}{%
\tikz[scale=0.23] {
    \draw[line width=0.7,line cap=round] (0,0) to [bend left=6] (1,1);
    \draw[line width=0.7,line cap=round] (0.2,0.95) to [bend right=3] (0.8,0.05);
}}
\newcommand{\tikzcmark}{%
\tikz[scale=0.23] {
    \draw[line width=0.7,line cap=round] (0.25,0) to [bend left=10] (1,1);
    \draw[line width=0.8,line cap=round] (0,0.35) to [bend right=1] (0.23,0);
}}
\newcommand{\preranker}{
OIE$^{\text{pre}}_{\text{ranker}}$
}
\newcommand{\reranker}{
Fact$^{\text{re}}_{\text{ranker}}$
}

\newcommand{\error}[1]{\small{$\pm$ #1}}

\definecolor{forestgreen}{rgb}{0.0, 0.27, 0.13}
\definecolor{forestgreen2}{rgb}{0.13, 0.55, 0.13}
\definecolor{green2}{rgb}{0.0, 0.65, 0.31}

\title{Linking Surface Facts to Large-Scale Knowledge Graphs}

\definecolor{myblue}{rgb}{0.9, 0.1, 0.94}

\newcommand*\circled[1]{\tikz[baseline=(char.base)]{
            \node[shape=circle,draw,inner sep=.6pt] (char) {#1};}}
            
\author{Gorjan Radevski$^{1,2}$, Kiril Gashteovski$^{1,3}$, Chia-Chien Hung$^1$, \\ \textbf{Carolin Lawrence$^1$, Goran Glavaš$^{4}$} \\ 
$^1$NEC Laboratories Europe, Heidelberg, Germany; $^2$KU Leuven, Leuven, Belgium; \\
$^3$CAIR, Ss. Cyril and Methodius University, Skopje, North Macedonia; \\ $^4$CAIDAS, University of Würzburg, Würzburg, Germany \\ 
\texttt{$^1$firstname.lastname@neclab.eu, $^2$gorjan.radevski@kuleuven.be,} \\ \texttt{$^4$goran.glavas@uni-wuerzburg.de}}

\begin{document}
\maketitle
\begin{abstract}

Open Information Extraction (OIE) methods extract facts from natural language text in the form of (\textit{``subject''}; \textit{``relation''}; \textit{``object''}) triples. These facts are, however, merely surface forms, the ambiguity of which impedes their downstream usage; e.g., the surface phrase \emph{``Michael Jordan''} may refer to either the former basketball player or the university professor. Knowledge Graphs (KGs), on the other hand, contain facts in a canonical (i.e., unambiguous) form, but their coverage is limited by a static schema (i.e., a fixed set of entities and predicates). To bridge this gap, we need the best of both worlds: (i) high coverage of free-text OIEs, and (ii) semantic precision (i.e., monosemy) of KGs. In order to achieve this goal, we propose a new benchmark with novel evaluation protocols that can, for example, measure fact linking performance on a granular triple slot level, while also measuring if a system has the ability to recognize that a surface form has no match in the existing KG. Our extensive evaluation of several baselines shows that detection of out-of-KG entities and predicates is more difficult than accurate linking to existing ones, thus calling for more research efforts on this difficult task. We publicly release all resources (data, benchmark and code)\footnote{\url{https://github.com/nec-research/fact-linking}}.

\end{abstract}

% Comments from Daniel for the abstract
% - reduce complexity of sentences in the beginning
% - go straight to the problem (e.g., drop "merely strings" part and go directly to ambiguity)
% - add finding at the end of abstract or phrase like "we hope to sparkle more research in this direction
% - drop term "slots", as it might not be clear to reviewers of what it refers to

\section{Introduction}

\begin{figure*}
    \centering
    \includegraphics[width=1.0\textwidth]{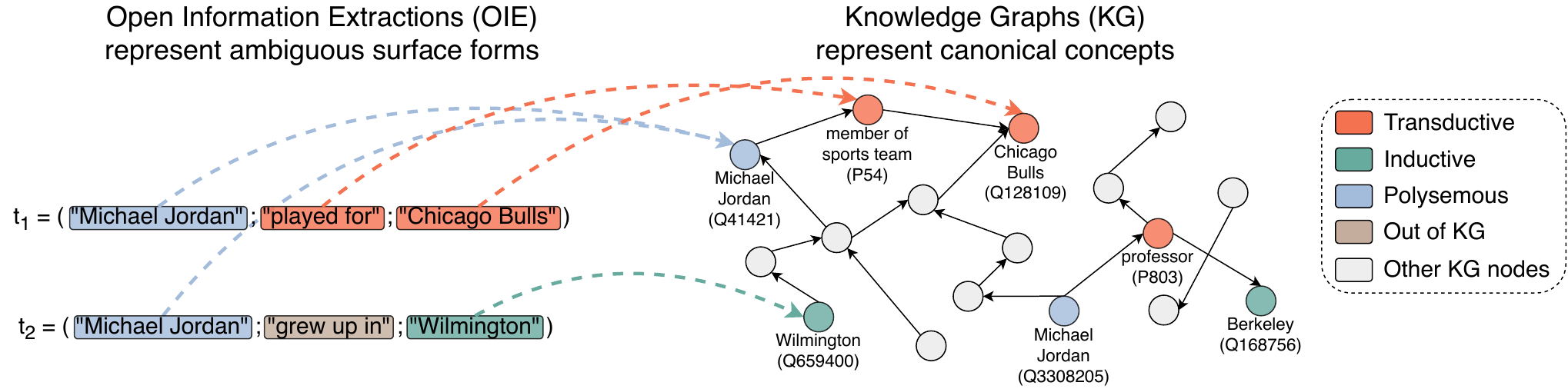}
    \caption{Open Information Extractions (OIEs) represent free-text surface-form triples, which may be ambiguous. Knowledge Graphs (KGs) represent canonical, unambiguous concepts, yet are limited by a hand-crafted schema. By linking OIEs to KG facts we bridge the gap between the schema-free (but ambiguous) surface facts extracted from text and the schema-fixed (but unambiguous)
    KG knowledge. Our benchmark addresses \textit{all} aspects of the OIE-to-KG linking. In \textcolor{transductive}{\textbf{transductive}} evaluation, the KG facts consist of entities \textit{seen} during training, while in \textcolor{inductive}{\textbf{inductive}} evaluation the KG facts contain entities \textit{strictly outside} of the training data. \textcolor{polysemous}{\textbf{Polysemous}} OIEs match to \textit{multiple} KG concepts (e.g., ``\emph{Michael Jordan}''). Finally, some OIEs might refer to \textcolor{outofkg}{\textbf{Out-of-KG}} entities or predicates.}
    % First, polysemous surface forms could match to more than one concept in the KG (``\emph{Michael Jordan}''). Second, transductive surface forms have been seen during training (``\emph{played for}''), whereas, third, inductive ones have not been seen  (``\emph{Wilmington}''). Finally, often surface forms cannot be linked to KGs because their concept is not present in the KG \cite{gashteovski2019opiec} (\emph{``grew up in''}). }
    % obtain the best of both worlds: high coverage of free-text OIEs and the semantic precision of KGs.}
    \label{fig:fig1}
\end{figure*}

%Open Information Extraction (OIE) methods extract surface \textit{("subject"; "relation"; "object")}-triples from natural language text, without requiring predefined schema \cite{Banko2007OpenIE}. Consider the following input sentence: \emph{``Michael Jordan, who grew up in Wilmington, played for Chicago Bulls''}; an OIE system should extract the following surface facts (i.e., OIE triples): $t_1$=\textit{("Michael Jordan"; "grew up in"; "Wilmington")} and $t_2$=\emph{("Michael Jordan"; "played for"; "Chicago Bulls")}. These structured representations of factual information are useful in various downstream tasks, including text summarization \cite{ribeiro-etal-2022-factgraph}, story comprehension \cite{andrus2022enhanced}, question answering \cite{wu2022triple}, event extraction \cite{chen2023led}, video grounding \cite{nan2021interventional}, and knowledge graph population \cite{gashteovski2020aligning}.

Open Information Extraction (OIE) methods extract surface \emph{(``subject''; ``relation''; ``object'')}-triples from natural language text in a schema-free manner \cite{Banko2007OpenIE}. For example, given the sentence \emph{``Michael Jordan, who grew up in Wilmington, played for Chicago Bulls''}, an OIE system should extract the triples (i.e., \emph{surface facts}): $t_1=$\emph{(``Michael Jordan''; ``played for''; ``Chicago Bulls'')} and $t_2=$\emph{(``Michael Jordan''; ``grew up in''; ``Wilmington'')}. The output of such systems is used in a diverse range of downstream tasks, including summarization \cite{Ribeiro2022FactGraphEF}, question answering (QA) \cite{wu2022triple}, event extraction \cite{dukic2023leveraging}, text clustering \cite{viswanathan2023large} and video grounding \cite{nan2021interventional}. 
%and knowledge graph population \cite{gashteovski2020aligning}.
%
%OIE triples, however, often contain ambiguous surface-form entities and relations. In the given examples from $t_1$ and $t_2$, the OIE entity mention \emph{``Michael Jordan''} can refer to different entities; e.g., the basketball player or the computer scientist. Resolving these ambiguities by linking OIE slots to clear and unambiguous concepts (e.g., entities and relations of a Knowledge Graph) is crucial. %The utilized KG information enhances the performance of downstream tasks, for instance, question answering \cite{saxena2020improving}, automatic medical diagnosing \cite{li2022neural}, or dialogue systems \cite{joko2021conversational}). %For a clearer illustration of the problem, see Fig.~\ref{fig:fig1}.
%Likewise, relation linking is important for downstream tasks such as ...
%
OIE triples, however, consist of surface-form entities and relations, which are frequently ambiguous (e.g., in $t_1$ and $t_2$, the entity mention \emph{``Michael Jordan''} may refer to several entities, e.g. the basketball player or the computer scientist). Resolving such ambiguities, by linking the OIE triple slots to inventories of unambiguous concepts, improves their downstream utility, e.g.,  
%in the performance of downstream tasks such 
in QA \cite{saxena2020improving}, automatic medical diagnosing \cite{li2022neural} or dialogue \cite{joko2021conversational}. %For a clearer illustration of the problem, see Fig.~\ref{fig:fig1}.
%Likewise, relation linking is important for downstream tasks such as 

Knowledge Graphs (KGs), on the other hand, are inventories of \emph{canonical} facts in the form of (subject; predicate; object)-triples, where each slot is a unique (i.e., unambiguous) concept \cite{vrandevcic2012wikidata}. 
%While the KGs structure information for real-world entities and the relations between them, 
KGs are, however, limited by their own static and often hand-crafted schema (i.e., fixed set of entities and predicates). As a consequence, methods that directly extract canonical KG triples from text \cite{distiawan2019neural, josifoski2021genie}, preemptively discard any information outside of the reference KG schema. Acknowledging this, \citet{ye2023schema} recently proposed \emph{schema-adaptable KG construction}, with the goal of extracting information for a KG with an evolving schema. \citet{ye2023schema} conclude that OIE methods indeed extract meaningful new knowledge for such KGs, however, they point precisely to the ambiguity of the surface forms as the major obstacle.  

To combine the best of both worlds, we need to bridge the gap between the schema-free (but ambiguous) surface facts extracted from text and the schema-fixed (but unambiguous) KG knowledge. However, existing benchmarks and models only partially address the problem. 
One line of work \cite{zhang2019openki, jiang2021cori, wood2021integrating} assumes a setting, which is arguably unrealistic, because the OIE entity slots are \textit{a priori} linked to KG entities and thus only addresses relation linking. %This is problematic, because the OIE entities are never linked apriori in realistic scenarios due to the ``open'' (i.e., \emph{surface}) nature of OIE systems. 
%%Other line of work \cite{distiawan2019neural, cabot2021rebel} neglects the OIE component and only provides a set of KG triples together with a text paragraph entailing them. 
A second line of research \cite{distiawan2019neural,cabot2021rebel,josifoski2021genie,sakor2020falcon,elsahar2018t} entails canonical KG triples directly from text paragraphs, bypassing OIE, which makes such models tied to the fixed KG schema. 
% Finally, a third line of work \cite{lin2020kbpearl, lin2021tenet, sakor2020falcon} evaluates OIE linking to a limited set of KG entities and predicates: ramification of unrealistic KG population benchmarks \cite{elsahar2018t, glass2018dataset}.
% resorts to standard KG population benchmarks, which are unrealistically small.
% \footnote{E.g., Wikidata \cite{vrandevcic2012wikidata} contains hundreds of millions of entities and thousands of predicates.}
%which come with an unrealistically small sets of entities and predicates
%, considering the massive size of real-world KGs today
%, such an assumption is unrealistic. 
Critically, \textit{all} these works conjecture that \emph{each} OIE slot \textit{has} a corresponding KG entry.  \citet{gashteovski2019opiec} showed that this is hardly ever the case in practice.

\rpar{Contributions and Findings} We move away from the unrealistic and incomplete assumptions of prior work and propose \circled{1} a novel large-scale benchmark for OIE-to-KG linking; 
%\circled{2} multifaceted evaluation protocols that cover all aspects of linking OIE facts to KGs (i.e., evaluating transductivity, inductivity, handling of polysemy, and out-of-KG detection); 
% \circled{2} multifaceted evaluation protocols that can measure the performance for the different aspects of linking OIE facts to KGs; 
\circled{2} multifaceted evaluation protocols that cover \textit{all} aspects of linking OIE facts to KGs (for an overview, see Fig.~\ref{fig:fig1});
\circled{3} several strong baselines, inspired by state-of-the-art entity linking \cite{wu2019scalable} and cross-modal retrieval \cite{miech2021thinking,geigle2022retrieve} approaches. 

%For an overview of the different facets in our benchmark see Figure \ref{fig:fig1}. Based on this, we find that current methods (i) perform well \textit{transductively}, where, at inference, KG facts consist of entities \textit{seen} during training, but (ii) their performance deteriorates in an \textit{inductive} evaluation, where inference KG facts contain entities outside of the training data. Further, we find that (iii) a dedicated OIE-to-KG fact re-ranking model improves the linking performance for polysemous OIEs and that (iv) we can obtain comparable performance by training models solely on synthetic data (i.e., with KG as the only human-annotated data). 

%We show an overview of the different facets in our benchmark in Fig.~\ref{fig:fig1}. 
Through our experimental study, we found that the methods (i) perform well \textit{transductively} but (ii) their performance deteriorates in an \textit{inductive} evaluation.
% Further, we find that (iii) a method to improve the linking performance for polysemous OIEs.
Further, we find that (iii) a dedicated OIE-to-KG fact re-ranking model improves the linking performance of both inductive and polysemous OIEs, and that (iv) we obtain high performance by training models solely on a synthetic variant of our dataset (i.e., with the KG as the only human-annotated data). 
% and that (iv) we can obtain comparable performance by training models solely on synthetic data (i.e., with KG as the only human-annotated data). 
%, while the rest is obtained automatically\footnote{By data processing pipelines, unsupervised OIE methods, Large Language Models, etc.}. 
%Lastly, we shed more light on the largely uninvestigated problem of detecting \textit{Out-of-Knowledge-Graph} extractions: we show that (iv) strong OIE slot linking models can be transformed into decent detectors of Out-of-KG entities (by means of an additional attention mechanism), but (v) the same does not hold for detection of out-of-KG predicates, a task that our experiments profile as a difficult open problem and one that requires much more research attention.
%Lastly, we shed more light on the largely uninvestigated problem of detecting \textit{Out-of-Knowledge-Graph} extractions. 
Lastly, we investigate the largely underexplored issue of detecting \textit{Out-of-Knowledge-Graph} extractions.
We show that (v) it is possible to detect Out-of-KG entities to an extent, however, the same does not hold for predicates: a task that our experiments identify as a difficult open problem, which 
requires more research attention.
\section{Fact Linking: Problem Statement}\label{sec:fl}
%We address the problem of linking open facts (OIE triples) to large-scale knowledge graphs (KGs). For example, given an OIE triple \emph{(``Michael Jordan''; ``played for''; ``Chicago Bulls'')}, the goal is to link each of the OIE slots to their canonical KG form; e.g., $t_{kg}$=(Q41421; P52; Q128109). In this example, Q41421, P19 and Q128109 refer to the concepts \emph{Michael Jordan (basketball player)}, \emph{member of sports team} and \emph{Chicago Bulls (NBA team)} respectively (Fig.~\ref{fig:fig1}). %In our setup, it is also possible for a given open relation or entity mention not to be present in the reference KG (e.g., the open relation \emph{"grew up in"} in Figure~\ref{fig:fig1}).

For a given surface-form OIE triple $t_1=(``s";``r";``o")$, the goal is to link each slot to a canonical concept in a KG (\textit{if} the corresponding concept exists in the KG): $``s" \rightarrow e_1 \in \mathcal{E}$; $``r" \rightarrow p \in \mathcal{P}$; $``o" \rightarrow e_2 \in \mathcal{E}$, with $\mathcal{E}$ and $\mathcal{P}$ as the (fixed) sets of KG entities and predicates. Importantly, our problem definition (and consequently evaluation) focuses on linking at the \textit{fact level}, where each OIE slot is contextualized with the other two OIE slots.\footnote{Alternatively, additional context can be included, e.g., the provenance from which the OIE surface fact is obtained.}

%which is particularly relevant for resolving ambiguous surface forms. 
%In particular, depending on the context of the OIE fact, the same surface form ``$s$'' can be mapped to several entities $\lbrace e_1, \ldots , e_n \rbrace$. 
%For these reasons, the linking is performed on \emph{fact instance level}. 
Additionally and crucially, we want linking models that can assign an empty set to OIE surface forms (e.g., $``s"$ $\rightarrow \emptyset$) when they refer to concepts not present in the KG (i.e., \textit{out-of-KG} concepts). 
%It is possible for a model to map a slot to an empty set , thus allowing for detecting out-of-KG concepts; i.e., entity or relation mentions that are outside of the predefined schema $KG$. 
%Figure \ref{fig:fig1} illustrates our multifaceted fact linking framework, with four different evaluation setups it supports. In what follows, we discuss each of these configurations (i.e., task formulations) in detail, including how they relate to prior work.
To enable a realistic setup for linking free-text OIE triples to a KG, we build a benchmark which considers four different facets (see Fig.~\ref{fig:fig1}).
%%%%% Assuming good coverage of RW in this section, not sure we need an extra RW section in the appenix
%\footnote{For more detailed discussion on related work, please refer to Appendix \ref{app:rel-work}.}.

%The problem of fact linking, however, has multiple facets (i.e., aspects). Namely, one can perform fact linking with 
%(1) \textit{transductive} evaluation tests models on linking of OIE facts that cover KG concepts (i.e., entities and relations) that the model has  were  that were already seen during training (i.e., transductive linking); (2) entity mentions that were not seen during training (i.e., inductive linking); (3) entity mentions that are polysemous (e.g., the mention \emph{``JFK''} could refer to both the former U.S. President and the airport in New York); or (4) option for yielding that an entity or relation slot is outside of the KG schema (i.e., out-of-KG detection). 
%
%.

\rpar{\textcolor{transductive}{Transductive Fact Linking}}
In transductive linking, we measure how well the models link OIEs to KG facts consisting of entities and predicates seen during training (as components of training KG facts). Note that the testing KG facts (as whole triples) are not in the training data. %to what extent the models can link OIE facts, whose slots refer to KG entity and relations observed during training.  
Consider, for example, the extraction $t_1$ in Fig.~\ref{fig:fig1}. In the transductive linking task, the mentions \emph{``played for''} and \emph{``Chicago Bulls''} refer to the KG predicate (P54) and entity (Q128109) respectively; both seen by the model as part of other training KG facts. However, the whole triple (Q41421; P54; Q128109) to which $t_1$ is linked was not part of the training data.

\rpar{\textcolor{inductive}{Inductive Fact Linking}}
%Complementary to the transductive setting, 
The inductive setup evaluates the linking to KG facts that consist of entities that are not seen during the training of the models. 
 In other words, this setup tests the generalization of fact linking models over entities. 
%Particularly, at inference time, the entity mentions that refer to KG entities are outside of the training data. 
% If we assume that the extraction $t_1$ from Figure \ref{fig:fig1} is  a test instance, then the inductive setup assumes that the KG entities Q41421 and Q128109 were not part of any training data KG triple.
In Fig.~\ref{fig:fig1}, given $t_2$ as a test instance, the OIE entity \emph{``Wilmington''} is inductive as it refers to a KG entity (Q659400) that is not part of any training KG fact.

\rpar{\textcolor{polysemous}{Polysemous Fact Linking}}
% In this setup,
We focus on OIEs for which the \textit{``s''} and \textit{``o''} slots are ambiguous w.r.t. the KG, i.e.,
% if considered
in isolation they refer to a set of KG entities rather than a single entity. The mention \emph{``Michael Jordan''} from either $t_1$ and $t_2$ (Fig.~\ref{fig:fig1}), in isolation, refers to both the basketball player (Q41421) and the computer scientist (Q3308205). Here, the other two OIE slots offer the disambiguation signal that is necessary for successful linking.
%disambiguate such links, the model has to rely on the context---i.e., the OIE as a whole---to perform successful 

\rpar{\textcolor{outofkg}{Out-of-KG Detection}} 
%are, by definition, of finite size of KGs Such assumption is rarely the case in practice due to the fixed-schema nature of KGs . 
We introduce a novel \textit{out-of-KG detection} task, in which the models are to recognize that an OIE triple component (i.e., \textit{``s''}, \textit{``r''} or \textit{``o''}) cannot be linked because they do not have a corresponding KG concept (e.g., the relation \emph{``grew up in''} from the triple $t_2$ in Fig.~\ref{fig:fig1}).
%recognize OIE triples that cannot be linked to the KG because at least one of the \textit{``s''}, \textit{``r''}, \textit{``o''} slots does not have a corresponding KG concept (e.g., the relation \emph{``grew up in''} from the triple $t_2$ in Fig.~\ref{fig:fig1}). 
%measure the Out-of-KG detection of OIE entities and relations. To do this, we select OIE $\leftrightarrow$ KG-fact pairs, where the KG-facts entities and relations cannot be found in the knowledge graph. 

%\subsection{Related Work}

\section{FaLB: \underline{Fa}ct \underline{L}inking \underline{B}enchmark}\label{sec:preparation}
We set up an automatic data processing pipeline to derive an OIE-to-KG fact linking benchmark, which supports all four evaluation facets from \S\ref{sec:fl}. We refer to both the process (i.e., pipeline) and the resulting benchmark as \textbf{FaLB}. FaLB's input is a dataset with (gold) alignments between natural language sentences and KG facts entailed by the sentence; i.e., each data instance is \emph{(sentence, KG fact)} pair. 
% FaLB's input is a dataset that contains text and corresponding (one or more) KG facts entailed by the sentence. Thus, each data point is a \emph{(sentence, KG fact)} pair. 
%FaLB starts from any dataset with (gold) alignments between natural language sentences, and entities and predicates from a reference KG that the sentences mention; i.e., each data point is \emph{(sentence, KG fact)} pair. 
Consequently, the creation of FaLB entails five design decisions: selection of (i) sentence-to-KG fact dataset(s), and (ii) a reference KG; (iii) generating OIE triples, (iv) high-precision OIE-KG fact alignments, and (v) a data augmentation strategy to increase the diversity of the data.
%selection of (i) a reference KG and (ii) OIE method(s), (iii) criterion for high-precision matching of OIE triples against KG facts, and (iv) a data augmentation strategy.
%that can be applied on any data samples which feature sentences as well as KG facts which are entailed in the sentences. To be specific, there are four main components in the pipeline: (i) A reference Knowledge Graph; (ii) OIE method(s); (iii) Matching criterion; (iv) Data Augmentation.
% See Figure~\ref{fig:process} for a visual depiction of the entire pipeline.
Below is an example instance of the FaLB dataset:

\begin{tcolorbox}[enhanced, fit to height = 4.2cm, colback = white, colframe = rowrow, boxsep = 2pt, coltitle = black, left = 1pt, right = 1pt, top = 1pt, bottom = 0pt, title = \textbf{Example Data Instance from FaLB }]
  \underline{\textbf{Sentence:}} \emph{``M.~J., who was born in Brooklyn, played for the Bulls.''} \vspace{2.5pt}
  
  \underline{\textbf{OIE surface facts:}} $t_1=$\emph{(``M.~J.''; ``played for''; ``the Bulls'')} and $t_2=$\emph{(``M.~J.''; ``was born in''; ``Brooklyn'')}\vspace{2.5pt}
  
  \underline{\textbf{KG canonical fact identifiers:}} (Q41421; P54; Q128109); (Q41421; P19; Q18419)\vspace{2.5pt}

  \underline{\textbf{KG text facts:}} (Michael Jordan; member of sports team; Chicago Bulls); (Michael Jordan; place of birth; Brooklyn)\vspace{2.5pt}
  
  \underline{\textbf{KG entity aliases:}} (Q41421: Air Jordan, Michael Jeffrey Jordan, His Airness); (Q128109: Bulls, The Bulls), ...
\end{tcolorbox}

\paragraph{Sentence-to-KG Fact Datasets.} We build FaLB benchmarks for two such existing datasets: REBEL \cite{cabot2021rebel} and SynthIE \cite{josifoski2023exploiting}. \textit{REBEL} is built from Wikipedia abstracts, where the manually hyperlinked entities in each sentence are linked to Wikidata entities, thus making them golden. To match the sentence with a KG fact, for each pair of KG entities $e_i$ and $e_j$ within the sentence, REBEL obtains all KG predicates $p_k$ such that $(e_i, p_k, e_j)$ or $(e_j, p_k, e_j)$ exist. However, two Wikidata entity nodes connected with a predicate (thus constituting a KG fact) do not guarantee the predicate validity in the sentence. For that reason, \citet{cabot2021rebel} use a Natural Language Inference pre-trained RoBERTa \cite{liu2019roberta} to filter the predicates \textit{not entailed} by the Wikipedia sentence; see \cite{cabot2021rebel} for details. We also use \textit{SynthIE}, which is synthetically generated using a Large Language Model (LLM) \cite{brown2020language}. Given a set of KG facts, the LLM is prompted to generate a sentence that covers the KG facts.\footnote{\citet{josifoski2023exploiting} manually evaluated that SynthIE is of higher quality than REBEL (see Appendix~\ref{app:rebel-vs-synthie} for details).}

\paragraph{Reference KG.}%I removed consutrction because that sounds like we created these KGs
We use Wikidata \cite{vrandevcic2012wikidata} as our reference KG, because REBEL and SynthIE align sentences to Wikidata facts. %\footnote{Wikidata is also one of the largest and most widely used publicly available KGs.} 
We select the subgraph of Wikidata that contains all entities which have a corresponding Wikipedia page, as per \citet{wu2019scalable}. 
%%
%and in particular the subset of the graph that  subset which contains a fixed set of entities $\mathcal{E}$ and relations $\mathcal{R}$. The subset of entities we keep is all Wikidata entries (Q-prefixed) which have a corresponding Wikipedia page associated with them as per \citet{wu2019scalable}. We further keep all Wikidata relations which appear as properties. 
Following \citet{lerer2019pytorch}, we additionally filter out the most infrequent entities and predicates, appearing less than 5 times in the whole Wikidata dump. This results in a large reference KG with $5,794,782$ unique entities and $4,153$ unique predicates.
%, which we denote as Large Knowledge Graph. 
We further create two smaller, dataset-specific reference KGs; one for each of the two datasets for which we apply FaLB: REBEL and SynthIE. These Benchmark-Restricted KGs (BRKGs) contain only the Wikidata entities and predicates referenced by at least one OIE triple extracted from their respective datasets. The BRKG for REBEL contains $625,125$ entities and $565$ predicates, while the one for SynthIE contains $702,334$ entities and $758$ predicates.
% REBEL: 625,125 entities; 565 relations.
% SynthIE: 702,334 entities; 758 relations.
% Full KG: 5.9M entities, 4k relations. In them are all entities/relations from REBEL and SynthIE.
% We apply all OIE methods on each sentence to obtain a set of open information extractions.
\paragraph{Generating OIE Triples.}
We use four state-of-the-art OIE methods to obtain a set of OIE triples for each of the sentences in the dataset. To increase diversity, we use two state-of-the-art rule-based OIE methods: MinIE \cite{gashteovski2017minie} and StanfordOIE \cite{angeli2015leveraging}; and two state-of-the-art neural OIE models: MilIE \cite{kotnis2022milie} and Multi$^2$OIE \cite{ro2020multi}.

%\paragraph{OIE-KG Fact Alignment in \textbf{FaLB}.}
\paragraph{High-precision OIE-KG Fact Alignments.}
% FaLB starts from any dataset with (gold) alignments between (i) natural language sentences and (ii) entities and predicates from a reference KG that those sentences mention.
% In this work, we build FaLB benchmarks for two such existing datasets: REBEL \cite{cabot2021rebel} and SynthIE \cite{josifoski2023exploiting}. 
%work, we build  that  align natural language sentences to from a corpus of sentences, each of which is associated with a set of KG t 
% We automatically extract OIE triples from each sentence with existing OIE systems and
Next, we need to match the extracted OIE triples against the set of KG facts associated with the sentences. Crucially, this automatic step needs to have a high-precision in order to create a high quality benchmark.
%We match the extracted OIE triples against the set of KG facts associated with the sentences. 
%%
% entailed in the sentence,
%%and a set of OIEs,
% obtained from the sentence,
%%we set up an OIE $\leftrightarrow$ KG-fact matching pipeline. 
%%
% The matching is done as follows.
As per the distant supervision assumption of \citet{mintz2009distant}, we create an alignment $t\leftrightarrow f$ between any OIE triple $t$ and any KG fact $f$ that exactly match (i.e., have identical text form) in subject and object, assuming that the relation of $t$ is aligned with the predicate of $f$.  
%for each OIE triple for which $s$ and $o$ are exact string matches with the subject and object of the KG triple  , has a subject and an object which are aligned (by exact string matching) with the surface form of the KG fact subject and object, we keep the OIE $\leftrightarrow$ KG-fact pair, and assume that the OIE predicate is aligned with the KG-fact relation (as per \citet{mintz2009distant} distant supervision assumption). 
When multiple OIE triples (excluding exact duplicates) $t_1$, $t_2$, \dots, $t_k$ (e.g., extracted with different OIE systems) match with the same KG fact $f$, we obtain all $k$ alignments: $t_1\leftrightarrow f$, \dots, $t_k\leftrightarrow f$. Finally, we remove all training alignments $t \leftrightarrow f$ that exist in the test portion. 
To verify that this strategy indeed has high precision,  
%To verify the data quality, 
we randomly sample 100 test instances, and conduct a human-study with two expert annotators. The annotators found 97\% correct pairings (inter-annotator agreement of 99\%; Kohen's kappa of 0.80); see Appendix~\ref{app:data-quality} for details. We therefore confirm the reliability of this design choice to produce high quality data.

\paragraph{Data Augmentation to Increase Diversity.}
Lack of example diversity---most linked OIE entity mentions exactly match the text of the corresponding KG entities---is a common problem in existing linking benchmarks \cite{cabot2021rebel}. This stems from strict data curation procedures that aim for high alignment precision, representing a mismatch with practice, where often the entity mentions do not exactly match the canonical KG entity denotation.
%m more frequenty  mismatch with texts , only a few entity mentions in text differ from their canonical KG from -- which is rarely the case in practice (refer to Figure~1). 
To train and evaluate fact linking methods on more complex linking examples (e.g., linking from initials and abbreviations), we augment the data using the additional information available in the reference KG. For each alignment $t \leftrightarrow f$, we fetch the Wikidata aliases (i.e., non-canonical denotations) for the subject and object entities of $f$. We then create additional pairs $t' \leftrightarrow f$ with augmented OIE triples $t'$ for all possible alias combinations. For example, for the original OIE triple \emph{(``Michael Jordan''; ``played for''; ``Chicago Bulls'')}, this process results in augmented triples such as \emph{(``Air Jordan''; ``played for''; ``Chicago Bulls'')}, \emph{(``M.J.''; ``played for''; ``The Bulls'')}, etc.
% For more details, see Appendix~\ref{app:entity-alias-augmentation}, where we extensively evaluate the impact of this particular design choice.
%\footnote{We did not apply the same augmentation strategy for the OIE relations, as these alias substitutions cannot be used as a drop-in replacement.}.

\section{OIE-to-KG Fact Linking Models}\label{sec:methods}

\begin{figure*}[t]
\centering
\resizebox{0.85\textwidth}{!}{
\includegraphics[width=1.0\textwidth]{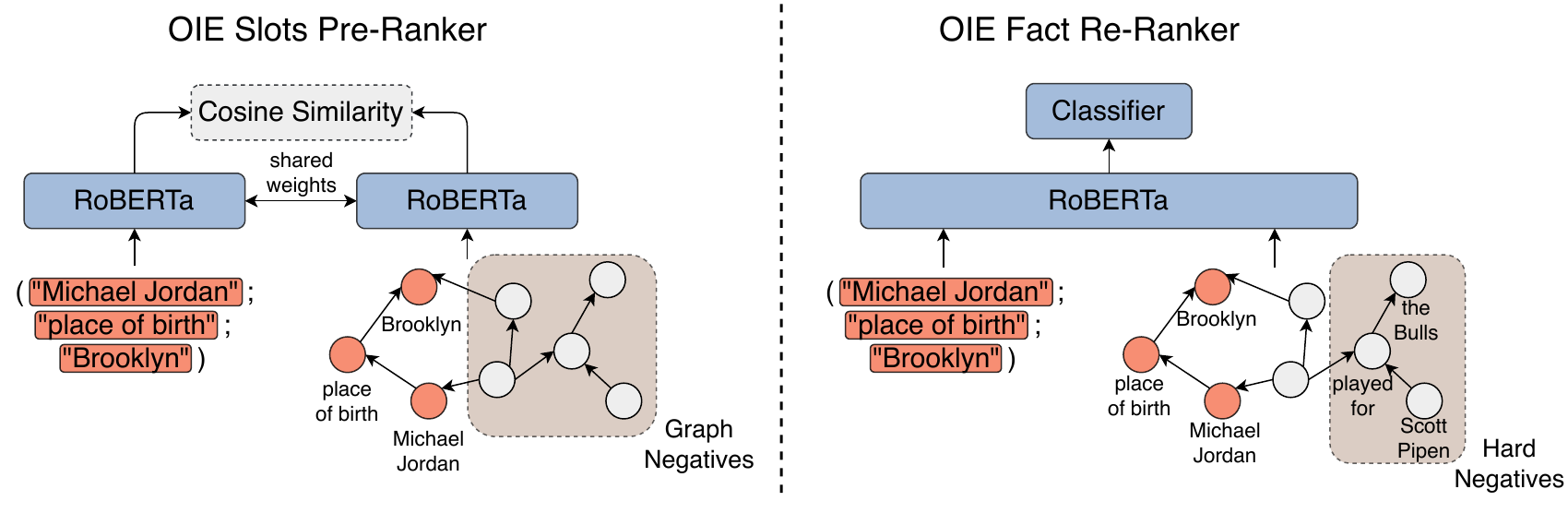}
}
\caption{
% An overview of the OIE linking models.
\textbf{Left:} \preranker which performs pre-ranking of the OIE slots to KG entries independently (i.e., no context between them is considered). Trained with negatives sampled from the whole KG; \textbf{Right:} \reranker which attends between the whole OIE triple and KG fact to output their similarity; Trained with hard-negatives.}
\label{fig:models}
\end{figure*}

Our goal is to link surface-form $(``s";``r";``o")$ OIE triples, to canonical Knowledge Graph facts $(e_1;p;e_2)$, where $e_1, e_2 \in \mathcal{E}$, and $p \in \mathcal{P}$.
% fixed set of entities $\mathcal{E}$ and relations $\mathcal{P}$ structured as [entity, relation entity] Knowledge Graph facts, where each
Each KG entity or predicate is represented as its surface-form KG label (e.g., ``Michael Jordan''), and its KG-provided description (e.g., ``American basketball player and businessman''). We henceforth refer to the entities and predicates as KG entries. Intuitively, since both data streams---the OIE triples and the KG facts---are in natural language, we opt to obtain their representations (i.e., embeddings) with a pre-trained language model. 
%\footnote{Notably, in this work we ignore the KG structure to obtain entity and predicate embeddings. Leveraging the underlying graph structure should yield entity and predicate representations more indicative of their semantics.}
We decouple the OIE-to-KG linking in two steps: pre-ranking and re-ranking (see \S\ref{sec:pre-rank} and \S\ref{sec:re-rank}, respectively),\footnote{We perform the linking in roughly $O(\vert\mathcal{E}\vert + \vert\mathcal{P}\vert + \vert\mathcal{E}\vert)$ time complexity, where $\vert\mathcal{E}\vert$ and $\vert\mathcal{P}\vert$ are the number of KG entities and predicates respectively. Directly encoding whole KG facts would have a prohibitive complexity: In the limit, we could have $\vert\mathcal{E}\vert \times \vert\mathcal{P}\vert \times \vert\mathcal{E}\vert$ KG facts.} as is done commonly in entity retrieval \cite{wu2019scalable} and image-text matching \cite{geigle2022retrieve,li2021align}.
See Fig.~\ref{fig:models} for an overview and Appendix~\ref{app:experimental} for implementation details.
%As encoding each of the KG facts is prohibitive,\footnote{In the limit, we could have $\vert\mathcal{E}\vert \times \vert\mathcal{P}\vert \times \vert\mathcal{E}\vert$ KG facts.} we decouple the OIE-to-KG linking in a pre-rank and re-rank phase (see \S\ref{sec:pre-rank} and \S\ref{sec:re-rank}, respectively).%: (i) \textit{OIE slot to KG entry} pre-ranking phase (\S\ref{sec:pre-rank}), followed by (ii) \textit{OIE triple to KG fact} re-ranking phase (\S\ref{sec:re-rank}); similarly used in entity retrieval \cite{wu2019scalable}, image-text matching \cite{geigle2022retrieve,li2021align}, etc. Given this formulation, we perform the linking in roughly $O(\vert\mathcal{E}\vert + \vert\mathcal{P}\vert + \vert\mathcal{E}\vert)$ time complexity, where $\vert\mathcal{E}\vert$ and $\vert\mathcal{P}\vert$ are the number of KG entities and predicates respectively.

\subsection{Pre-ranking OIE Slots to KG Entries}\label{sec:pre-rank}
We denote this model as \preranker. It aims to generate OIE slot embeddings and KG entry embeddings, such that an OIE slot embedding yields higher similarity with its aligned KG entry's embedding compared to the other KG entries.
%An OIE slot embedding yields higher similarity with its aligned KG entry's embedding than with the other KG entries.
Therefore, during training, we contrast the positive pairs against a set of negatives, thereby training the model to generate embeddings for a matching OIE slot and KG entry that lie close in the latent space. The motivation for such formulation is two-fold: (i) the number of entities is large, and could practically grow further, therefore computing the softmax over all KG entities during training is prohibitive; (ii) there may be unseen KG entries that we encounter during inference, therefore posing the problem as a standard classification inevitably leads to the model ignoring them. We use RoBERTa \cite{liu2019roberta} to encode the OIE slots and the KG entries.

\paragraph{OIE Embeddings.}
% To obtain an OIE embedding,
We first add special tokens to indicate the start of each OIE slot: \texttt{<SUBJ>} for \emph{``subject''}, \texttt{<REL>} for \emph{``relation''}, \texttt{<OBJ>} for \emph{``object''}. Hence, $t_1$ = \emph{(``M. Jordan''; ``grew up in''; ``Wilmington''}) is represented as \emph{``\texttt{<SUBJ>} M. Jordan \texttt{<REL>} grew up in \texttt{<OBJ>} Wilmington''}. We then tokenize the OIE representation, denoted as $\hat{t}$, and obtain RoBERTa token embeddings. Finally, we pool \textit{only} the special tokens' embeddings subsequently linearly projected in the desired latent space: $\hat{o}_{i} = \text{Linear}(\text{RoBERTa}(\hat{t}))$, where $\hat{o}$ is the slot embedding, and $i \in\mathbb{R}^3$ is the OIE slot index.

\paragraph{KG Embeddings.}
% To obtain KG embeddings,
We represent both the entities and predicates as their label followed by their description (if available in the KG); e.g., the entity ($e$) \textit{Michael Jordan} is represented as ``Michael Jordan \texttt{<DESC>} American basketball player and businessman...'', and the predicate ($p$) \textit{place of birth} is represented as ``place of birth \texttt{<DESC>} most specific known birth location of a person...''. The \texttt{<DESC>} special token indicates the start of the description. We then tokenize the representation ($\hat{e}$ for entity, $\hat{p}$ for predicate), and obtain embeddings using the same RoBERTa model. We pool the \texttt{<CLS>} token embedding and linearly project it in the OIE representation space as: $\hat{k}_{j} = \text{Linear}(\text{RoBERTa}(b_j))$, where $j \in \{\hat{e}, \hat{p}\}$, $\hat{k}$ is the KG entry embedding, and $b$ is the tokenized KG entry representation.
% Note that the embeddings of the knowledge graph are computed offline and stored on disk, i.e., linking an OIE to the knowledge graph requires obtaining embeddings of the OIE slots followed by a dot-product similarity with the knowledge graph entity or relation embeddings.\par

\paragraph{Linking OIEs to KG Facts.} Given an OIE slot embedding $o_i$ and a KG entry embedding $k_j$, we obtain their dot-product as: $\hat{s}_{pre} = o_i^T k_j$, where $o_i$ and $k_j$ are norm-scaled (i.e., $\hat{s}_{pre}$ represents the \preranker cosine similarity).
% $\hat{s} = \frac{o_i^T k_j}{\lVert o_i \rVert \lVert k_j \rVert}$,
% where $\hat{s}$ is the cosine similarity.
During inference, we link an OIE to a KG fact by selecting the most similar KG entry for each of the OIE slots.

\paragraph{\preranker Training.}\label{paragraph:preranker-training}
We train the model using standard contrastive loss: we sample $N-1$ in-batch negative KG entries for each positive OIE slot $\leftrightarrow$ KG entry pair, where $N$ is the batch size. As per standard practice \cite{oord2018representation}, we train the model using temperature-scaled InfoNCE contrastive loss: $\mathcal{L} = -\operatorname{log}\frac{e^{\hat{o}^\text{T}\hat{k}/\tau}}{e^{\hat{o}^\text{T}\hat{k}/\tau} + \sum_{n=1}^{\text{N}-1} e^{\hat{o}^\text{T}\hat{k}_n^{-}/\tau}}$, where $\tau$ is the temperature. Note that during training, we only sample negative KG entry embeddings for each OIE slot, but not the other way around.

% \paragraph{Negative Sampling From Whole KG.}
Sampling only in-batch negatives presents an issue as the training data represents only a limited subset of the whole KG (i.e., only the KG entries with paired OIE).
% s $\leftrightarrow$ KG aligned data using FaLB).
% However, the negative entities \& relations the model observes during training represent only a limited subset of the knowledge graph, that is, \textit{only} the subset of entities \& relation present in the dataset.
During inference, however, we contrast each OIE slot against the whole KG to find the KG entry with which it exhibits the highest similarity.
% , we contrast it against the whole KG -- presenting a discrepancy between training and testing.
% , and returning the KG element with the highest similarity.
Therefore, for each OIE slot, we additionally sample $e$ negative entities and $p$ negative predicates at random from the whole KG (e.g., we would sample $\sum_{n=1}^{\text{N}-1 + p} \hat{k}_n^{-}$ negative predicates).

%\paragraph{Negative Sampling From Whole KG.}
%Sampling only in-batch negatives presents an issue as the KG entries appearing in the training data represent only a limited subset of the whole KG (i.e., only the ones with paired OIE).
% s $\leftrightarrow$ KG aligned data using FaLB).
% However, the negative entities \& relations the model observes during training represent only a limited subset of the knowledge graph, that is, \textit{only} the subset of entities \& relation present in the dataset.
%During inference however, to find the KG entry with which the OIE slot exhibits the highest similarity, we contrast it against the whole KG -- presenting a discrepancy between training and testing.
% , and returning the KG element with the highest similarity.
%Therefore, besides the in-batch negatives, for each OIE slot, we sample $e$ negative entities and $p$ negative predicates at random from the whole KG (e.g., we would sample $\sum_{n=1}^{\text{N}-1 + p} \hat{k}_n^{-}$ negative predicates). In the experiments, whenever we use negatives from the whole KG in addition to the in-batch negatives, we abbreviate as $+ \text{GN}$.
% We denote this model as Slots pre-ranker with graph negatives (GN). This way, during training, each OIE slot $\leftrightarrow$ KG entity/relation pair gets contrasted against any of the graph entities/relations.

\subsection{Re-ranking OIEs to KG Facts}\label{sec:re-rank}
% In certain scenarios, e.g., when dealing with a polysemous OIE, linking to the appropriate KG fact is difficult without considering the context of the complete KG fact.
In certain scenarios (e.g., in the case of polysemous OIEs), matching whole OIEs with whole KG facts (i.e., not decoupled per OIE slot) could resolve the ambiguity and thus improve performance. To that end, for each OIE slot, we re-rank the \preranker top-k most probable KG links. We denote this model as \reranker. We perform self-attention between the OIE and the KG fact (both provided as input, separated by a \texttt{<FACT>} special token) with a single RoBERTa transformer, and return their similarity as:
% and, compared to the \preranker, which independently generates embeddings of the OIE slots and the knowledge graph entities \& relations,
% the \reranker relies on the attention mechanism to measure the OIE $\leftrightarrow$ KG fact similarity:
$\hat{s}_{re} = \sigma(\text{Linear}(\text{RoBERTa}(\hat{c})))$,  
% which are returned from the pre-ranking model. Namely, the input to the re-ranker is the concatenated OIE and knowledge graph fact proposal, while the output is the matching score between them normalized between 0-1:
where $\hat{c}$ is the concatenated OIE and KG fact representation, and $\hat{s}_{re}$ is the sigmoid ($\sigma$) normalized
% re-ranker OIE-to-KG fact
similarity.
% between the OIE and the KG Fact.
% We add an additional special token \texttt{<FACT>} between the OIE and the KG facts before providing them as input to the RoBERTa model.
% Additionally, we randomly corrupt the representation where we entirely drop the description of the KG fact slots with 50\% probability.
%\paragraph{Training Re-ranker}
%We train the re-ranker by sampling positives (i.e., a matching OIE $\leftrightarrow$ KG fact pairs) and negatives, where we corrupt some of the KG fact slots (subject, predicate, or object) by replacing them with incorrect ones. A straightforward approach is to corrupt the KG fact slots with KG entries sampled at \textit{random}. However, such model easily distinguishes between the correct and incorrect pairs, and yields poor re-rankings when prompted with the top-k \preranker links. To overcome the issue, we train  \reranker by sampling hard negatives.
%\paragraph{Generating Hard Negatives.}
%We first obtain embeddings for each KG entry using the \preranker, and find its top-k most similar candidates by measuring the similarity w.r.t. all other KG entries. We then corrupt the ground truth KG fact by randomly sampling \textit{only} from the top-k (hard) negative candidates. Lastly, with 50\% probability, we randomly mask (i.e., replace with a \texttt{<mask>} token) the description of the KG fact entries.

\paragraph{\reranker Training.}
We train by sampling matching OIE $\leftrightarrow$ KG fact pairs as positives, and negatives, where we replace some KG fact slots (subject, predicate, object) with incorrect ones, generated as follows:
% A straightforward approach is to corrupt the KG fact slots with KG entries sampled at \textit{random}. However, such model easily distinguishes between the correct and incorrect pairs, and yields poor re-rankings when prompted with the top-k \preranker links. To overcome the issue, we train  \reranker by sampling hard negatives.
We first obtain embeddings for each KG entry using the \preranker, and find its top-k most similar candidates w.r.t. all other KG entries. We then corrupt the ground truth KG fact by randomly sampling \textit{only} from the top-k (hard) negative candidates. Lastly, with 50\% probability, we randomly mask (i.e., replace with a \texttt{<mask>} token) the description of the KG fact entries.
\section{Experiments and Discussions}
We measure accuracy to evaluate both OIE slot linking to KGs (\S\ref{sec:kg-population}), and Out-of-KG detection of OIE slots (\S\ref{sec:out-of-kg-detection}).
% We use accuracy to evaluate both the OIE linking (per-each slot) and Out-of-KG detection.
To measure OIE fact linking, we score a hit if \emph{all} OIE slots are linked correctly. The error bars represent the standard error of the mean.
% INFO:root:subj = 77.8 +/- 0.08
% INFO:root:rel = 92.22 +/- 0.05
% INFO:root:obj = 92.24 +/- 0.05
% INFO:root:fact = 70.24 +/- 0.09

\begin{table*}[t]
\centering
\resizebox{0.9\textwidth}{!}{
\begin{tabular}{lccccccccccc} \toprule
    \rowcolor{rowrow} {Method} & {Split Type} & {Subject} & {Relation} & {Object} & {Fact} & {Subject} & {Relation} & {Object} & {Fact} \\ \midrule
    \rowcolor{row} {} & {} & \multicolumn{4}{c}{\textit{Benchmark-Restricted Knowledge Graph}} & \multicolumn{4}{c}{\textit{Large Knowledge Graph}} \\ \midrule
    Random & Transductive & 0.0 \error{0.0} & 0.0 \error{0.0} & 0.0 \error{0.0} & 0.0 \error{0.0} & 0.0 \error{0.0} & 0.0 \error{0.0} & 0.0 \error{0.0} & 0.0 \error{0.0} \\
    Frequency & Transductive & 0.0 \error{0.0} & 53.3 \error{0.1} & 8.1 \error{0.1} & 0.0 \error{0.0} & 0.0 \error{0.0} & 53.3 \error{0.1} & 8.1 \error{0.1} & 0.0 \error{0.0} \\
    SimCSE & Transductive & 5.4 \error{0.1} & 0.0 \error{0.0} & 0.9 \error{0.1} & 0.0 \error{0.0} & 2.6 \error{0.1} & 0.0 \error{0.0} & 0.5 \error{0.0} & 0.0 \error{0.0} \\
    % \preranker (no GN) & Transductive & 81.0 \error{0.1} & 91.5 \error{0.1} & 93.5 \error{0.1} & 71.3 \error{0.1} & 69.8 \error{0.2} & 89.6 \error{0.1} & 88.7 \error{0.1} & 58.4 \error{0.2} \\
    \preranker & Transductive & \textbf{86.8 \error{0.1}} & \textbf{93.5 \error{0.1}} & \textbf{95.7 \error{0.0}} & \textbf{79.1 \error{0.1}} & \textbf{78.7 \error{0.2}} & \textbf{93.5 \error{0.1}} & \textbf{93.1 \error{0.1}} & \textbf{70.7 \error{0.2}} \\
    $+$ Context & Transductive & 84.9 \error{0.1} & 92.2 \error{0.1} & 94.8 \error{0.1} & 77.7 \error{0.1} & 77.8 \error{0.1} & 92.2 \error{0.1} & 92.2 \error{0.1} & 70.2 \error{0.1} \\ \midrule
    Frequency & Inductive & 0.0 \error{0.0} & 0.1 \error{0.0} & 0.0 \error{0.0} & 0.0 \error{0.0} & 0.0 \error{0.0} & 0.1 \error{0.0} & 0.0 \error{0.0} & 0.0 \error{0.0} \\
    SimCSE & Inductive & 12.6 \error{0.3} & 0.4 \error{0.1} & 6.8 \error{0.3} & 0.0 \error{0.0} & 8.1 \error{0.3} & 0.0 \error{0.0} & 3.2 \error{0.2} & 0.0 \error{0.0} \\
    % \preranker (no GN) & Inductive & 69.4 \error{0.5} & 61.8 \error{0.5} & 58.2 \error{0.5} & 30.6 \error{0.5} & 58.2 \error{0.5} & 59.5 \error{0.5} & 46.3 \error{0.5} & 21.3 \error{0.4} \\
    \preranker & Inductive & 71.9 \error{0.4} & 69.8 \error{0.5} & 59.5 \error{0.5} & 34.9 \error{0.5} & 62.0 \error{0.5} & 69.8 \error{0.5} & 48.2 \error{0.5} & 25.4 \error{0.4} \\
    % Slots pre-ranker (GN) & SynthIE & REBEL & Inductive & \textbf{65.67 \error{0.50}} & 55.36 \error{0.52} & \textbf{55.60 \error{0.52}} & 26.75 \error{0.46} \\
    $+$ Context & Inductive & 74.5 \error{0.4} & \textbf{73.8 \error{0.4}} & \textbf{62.3 \error{0.5}} & 38.9 \error{0.5} & \textbf{64.5 \error{0.5}} & \textbf{73.8 \error{0.4}} & 50.8 \error{0.5} & 29.2 \error{0.5} \\
    $+$ \reranker & Inductive & \textbf{76.2 \error{0.4}} & 67.6 \error{0.5} & 60.8 \error{0.5} & \textbf{40.6 \error{0.5}} & \textbf{64.8 \error{0.5}} & 66.5 \error{0.5} & \textbf{54.8 \error{0.5}} & \textbf{32.9 \error{0.5}} \\ \midrule
    Frequency & Polysemous & 0.0 \error{0.0} & 68.1 \error{0.4} & 11.7 \error{0.3} & 0.0 \error{0.0} & 0.0 \error{0.0} & 68.1 \error{0.4} & 11.7 \error{0.3} & 0.0 \error{0.0} \\
    SimCSE & Polysemous & 1.3 \error{0.1} & 0.0 \error{0.0} & 0.3 \error{0.1} & 0.0 \error{0.0} & 0.4 \error{0.1} & 0.0 \error{0.0} & 0.0 \error{0.0} & 0.0 \error{0.0} \\
    % \preranker (no GN) & Polysemous & 64.0 \error{0.4} & 90.6 \error{0.2} & 90.9 \error{0.2} & 55.2 \error{0.4} & 49.6 \error{0.4} & 89.2 \error{0.3} & 86.7 \error{0.3} & 40.5 \error{0.4} \\
    \preranker & Polysemous & 68.8 \error{0.4} & 93.0 \error{0.2} & 94.6 \error{0.2} & 62.5 \error{0.4} & 58.2 \error{0.4} &  93.0 \error{0.2} & 92.4 \error{0.2} & 51.7 \error{0.4} \\
    $+$ Context & Polysemous & \textbf{77.9 \error{0.3}} & \textbf{93.3 \error{0.2}} & \textbf{95.8 \error{0.2}} & \textbf{71.6 \error{0.4}} & 69.5 \error{0.4} & \textbf{93.3 \error{0.2}} & 94.0 \error{0.2} & 63.5 \error{0.4} \\
    %$+$ \reranker & Polysemous & \textbf{73.9 \error{0.4}} & 90.0 \error{0.2} & 94.9 \error{0.2} & \textbf{67.7 \error{0.4}} & \textbf{68.8 \error{0.4}} & 89.8 \error{0.3} & 94.4 \error{0.2} & 63.0 \error{0.4} \\
    $+$ \reranker & Polysemous & 75.3 \error{0.4} & 90.6 \error{0.2} & 95.3 \error{0.2} & 69.5 \error{0.4} & \textbf{74.1 \error{0.4}} & 90.6 \error{0.2} & \textbf{95.6 \error{0.2}} & \textbf{69.3 \error{0.4}} \\ \bottomrule
\end{tabular}
}
\caption{OIE slot and fact linking accuracy with models trained and evaluated on REBEL. Evaluation on a smaller Benchmark-Restricted KG, and a Large KG. The error bars indicate standard error of the mean.}
\label{table:all-splits-new}
\end{table*}

% We first explore the models' ability to perform OIE slot linking to KGs (\S\ref{sec:kg-population}). Here, we focus on how the models deal with KG facts consisting of entities which exhibit specific properties -- transductive, inductive and polysemous.
% Secondly, we explicitly explore to what extent we are able to detect Out-of-Knowledge graph OIEs (\S\ref{sec:out-of-kg-detection}).

\subsection{Linking OIEs to Knowledge Graphs}\label{sec:kg-population}
\paragraph{Setup.} We explore the extent to which we can link OIE slots to a large-scale KG (Wikidata).
% We use the models described in \S\ref{sec:methods} (including several baselines), and
We address two main research questions concerning the OIE-to-KG fact linking task: (i) To what extent do methods generalize to different KG entity facets? We consider transductive, inductive, or polysemous entities (see \S\ref{sec:fl} for detailed definition); (ii) What is the performance impact of the KG size? We test two reference KG sizes: Benchmark Restricted KG ($\sim$650k entities, $\sim$0.6k predicates) and Large KG ($\sim$5.9M entities, $\sim$4k predicates).

\paragraph{Methods.} We use the following methods to obtain results for the OIE slot linking task: (i) \textsc{Random:} for each OIE slot, we sample a random KG entry; (ii) \textsc{Frequency:} based on the training data statistics, we link each OIE slot to the most frequent KG entry (entity or predicate) in the training set; (iii) \textsc{SimCSE:} we use a pretrained SimCSE model \cite{gao2021simcse}, where we represent each OIE slot and each KG entry (optionally, its description as well -- if available in the KG) in a natural language format, and obtain their embeddings. We
% \footnote{Given the entity Michael Jordan, we structure the input as: ``Michael Jordan (American basketball player and businessman (born 1963))''},
finally obtain the cosine similarity between each OIE slot and KG entry to perform the linking; (iv) \preranker: We train a pre-ranking model as per the setup described in \S\ref{paragraph:preranker-training};
(v) $+$~Context: We append the context sentence to the OIE\footnote{We structure the input as: ``OIE \texttt{<SENT>} Sentence'', where \texttt{<SENT>} is a special separator token.};
% (v) \preranker$+\text{GN}$: We also train with negatives sampled from the entire KG;
(vi) $+$\reranker: We additionally re-rank the top-k ($k=3$) pre-ranked OIE slot links as per the setup in \S\ref{sec:re-rank}.
%\footnote{We exclude entity linking methods \cite{wu2019scalable,decao2021autoregressive} from the analysis, as they most likely encountered much of our testing data during training.}

\paragraph{Results.}
In Table~\ref{table:all-splits-new} we report the OIE-to-KG linking performance for each OIE slot, as well as linking on fact level. Overall, across the data splits, we observe that all unsupervised baselines perform poorly compared to models proposed in this work; indicating that OIE linking is not trivial, hence off-the-shelf zero-shot models fail.\footnote{Note that simply linking the OIE relation to the most frequent KG predicate yields high accuracy, due to the KG predicates imbalance in REBEL.}
% Further, training with negatives sampled from the entire KG ($+ \text{GN}$) increases performance across all data splits. This is expected, as sampling KG negatives allows us to contrast the positive pair against a more diverse set of negatives.
Additionally, there is a significant performance drop on the inductive and polysemous split compared to the transductive split, suggesting that the models are neither robust w.r.t.~entities unseen during training, nor can cope with polysemous entities. Expectedly, leveraging extra context (via the sentence from where the OIE is obtained) aids the linking process in the inductive and polysemous split, as it helps generalization and disambiguation. Similarly, the \reranker brings a significant performance gain especially prominent for linking complete facts.
% Intuitively, polysemous entities are referred to by OIE slots with the same text, and due to ambiguity it is difficult to link them individually. -- Gorjan: This was in the original version (reviewed), but altered the text a bit based on the updated results (with sentence-context)
Finally, we observe a significant performance impact of the KG size: across all splits, OIE-to-KG linking is more challenging on large-scale KG compared to the smaller Benchmark-Restricted KG.
\paragraph{Training on Synthetic Data Improves Performance.}
We explore the extent to which we can learn fact linking models using synthetic data. SynthIE \cite{josifoski2023exploiting} is a dataset that features natural language sentences paired with KG facts, where the sentences are obtained using a LLM. Namely, given a set of KG facts, \citet{josifoski2023exploiting} prompt the LLM to generate a sentence which mentions (i.e., entails) all of the KG facts. Notably, if we can link OIE slots to KG facts by training on such synthetic dataset, then the KG remains the only human-annotated component for learning the OIE-to-KG fact linking task.
% -- therefore reducing the potential costs for creating such datasets.
% \textcolor{red}{This makes such methods useful for reducing the potential costs for creating such datasets.}

To measure to what extent we can link OIEs to KG facts using only synthetic data, we train \preranker models on both REBEL and SynthIE, and report results (in Table~\ref{table:across-splits}) on inductive testing splits w.r.t. each dataset.\footnote{To preserve the inductive property, SynthIE's inductive test split contains only entities found in SynthIE that are not present in the REBEL train data.}
\begin{table}[t]
\centering
\resizebox{1.0\columnwidth}{!}{
\begin{tabular}{lccccc} \toprule
    % \rowcolor{rowrow} {} & {} & \multicolumn{4}{c}{Benchmark Index} & \multicolumn{4}{c}{Entire Wikidata Index} \\
    \rowcolor{rowrow} {Train Data} & {Test Data} & {Subject} & {Relation} & {Object} & {Fact} \\ \midrule
    REBEL & REBEL & 62.0 \error{0.5} & 69.8 \error{0.5} & 48.2 \error{0.5} & 25.4 \error{0.4} \\
    REBEL & SynthIE & 53.6 \error{0.4} & 57.2 \error{0.4} & 44.9 \error{0.4} & 17.4 \error{0.3} \\
    \multicolumn{2}{l}{\large\textbf{\textit{Macro Score}}} & 57.8 & 63.5 & 46.6 & 21.4 \\ \midrule \midrule
    SynthIE & REBEL & 69.7 \error{0.3} & 63.2 \error{0.3} & 41.6 \error{0.3} & 22.4 \error{0.2} \\
    SynthIE & SynthIE & 64.1 \error{0.6} & 81.2 \error{0.5} & 57.6 \error{0.6} & 34.5 \error{0.6} \\
    \multicolumn{2}{l}{\large\textbf{\textit{Macro Score}}} & \textbf{66.9} & \textbf{72.2} & \textbf{49.6} & \textbf{28.5} \\
    % SynthIE & REBEL & 63.85 \error{0.50} & 54.36 \error{0.52} & 44.89 \error{0.37} & X \\
    % SynthIE & REBEL & 65.67 \error{0.50} & 55.36 \error{0.52} & 55.60 \error{0.52} & X \\
    % PR + GN w/ FRR & REBEL & SynthIE & 52.33 \error{0.37} & 52.71 \error{0.37} & 46.18 \error{0.37} \\ 
    \bottomrule
\end{tabular}
}
\caption{Experiments on human-created (REBEL) and synthetically generated (SynthIE) datasets. OIE linking to a Large KG variant on inductive splits w.r.t. each dataset (see Appendix~\ref{app:inductive-splits} for details).
% Surprisingly, using synthetic training data (SynthIE) outperforms using human-created training data on the human-created test set. This is an important insight for practical applications: to enable fact linking we no longer necessarily need a human-created training set.
}
\label{table:across-splits}
\end{table}
We observe that models trained on SynthIE are overall better OIE-to-KG fact linkers than models trained on REBEL (i.e., higher macro accuracy across datasets). This indicates that learning to link OIEs to KGs is possible using only synthetic data, thus the only human-annotated requirement remains to be a reference KG.

\paragraph{Ablation Study: Importance of Entity Alias Augmentation.} We observe that in current datasets most surface form entities (in the natural language sentences) appear only with their ``canonical'' label.\footnote{REBEL is constructed from Wikipedia abstracts, where the references use the canonical form name; SynthIE provides the KG fact (in text format) as is to the LLM, so naturally, the sentence generated does not feature the entity aliases. Therefore, \emph{Michael Jordan}'s synonyms such as \emph{M.J., Air Jordan} and \emph{His Airness,} rarely appear in the data.}
% This presents an issue as it diverges from how people refer to entities in natural language (i.e., using abbreviation, synonyms, etc.).
Since the OIEs represent surface-form facts, such lack of diversity prevents the models from learning more complex linking patterns. To overcome this, we perform entity alias augmentation in \textbf{FaLB} by adding the surface form aliases of the entities---available in Wikidata and manually curated---and ablate its impact on the OIE linking task. Besides \preranker models trained on REBEL and SynthIE \textit{with} entity alias augmentation, we train additional models \textit{without} the alias augmented samples. We report results in Table~\ref{table:alias-augment} on inductive REBEL and SynthIE testing data, which \textit{does} and \textit{does not} feature entity aliases.

\begin{table}[t]
\centering
\resizebox{1.0\columnwidth}{!}{
\begin{tabular}{cccccc} \toprule
    \rowcolor{rowrow} {Training} & {Testing} & {Subject} & {Relation} & {Object} & {Fact} \\
    \rowcolor{rowrow} \multicolumn{2}{c}{Augmentation} & \multicolumn{4}{c}{} \\ \midrule
    \rowcolor{row} \multicolumn{2}{c}{} & \multicolumn{4}{c}{\textbf{\textit{Models trained and evaluated on \underline{REBEL}}}} \\ \midrule
    \tikzxmark & \tikzxmark & 90.7 \error{0.1} & 91.9 \error{0.1} & 87.1 \error{0.2} & 74.0 \error{0.2} \\
    \tikzxmark & \tikzcmark & 61.6 \error{0.2} & 64.8 \error{0.2} & 41.5 \error{0.2} & 25.2 \error{0.2} \\ 
    \multicolumn{2}{l}{\textbf{\textit{Macro Score}}} & 76.2 & 78.4 & 64.3 & 49.6 \\ \midrule
    \tikzcmark & \tikzxmark & 89.0 \error{0.1} & 92.8 \error{0.1} & 85.1 \error{0.2} & 72.4 \error{0.2} \\ 
    \tikzcmark & \tikzcmark & 79.0 \error{0.2} & 92.2 \error{0.1} & 89.0 \error{0.1} & 67.9 \error{0.2} \\
    \multicolumn{2}{l}{\textbf{\textit{Macro Score}}} & \textbf{84.0} & \textbf{92.5} & \textbf{87.1} & \textbf{70.15} \\ \midrule \midrule
    \rowcolor{row} \multicolumn{2}{c}{} & \multicolumn{4}{c}{\textbf{\textit{Models trained and evaluated on \underline{SynthIE}}}} \\ \midrule
    \tikzxmark & \tikzxmark & 91.8 \error{0.2} & 86.3 \error{0.2} & 90.7 \error{0.2} & 72.8 \error{0.3} \\
    \tikzxmark & \tikzcmark & 61.2 \error{0.2} & 70.1 \error{0.2} & 40.7 \error{0.2} & 22.6 \error{0.2} \\
    \multicolumn{2}{l}{\textbf{\textit{Macro Score}}} & 76.5 & 78.2 & 65.7 & 47.7 \\ \midrule
    \tikzcmark & \tikzxmark & 91.1 \error{0.2} & 88.9 \error{0.2} & 90.5 \error{0.2} & 74.3 \error{0.3} \\
    \tikzcmark & \tikzcmark & 80.1 \error{0.2} & 89.9 \error{0.1} & 86.3 \error{0.2} & 64.8 \error{0.2} \\
    \multicolumn{2}{l}{\textbf{\textit{Macro Score}}} & \textbf{85.6} & \textbf{89.4} & \textbf{88.4} & \textbf{69.6} \\ \bottomrule
\end{tabular}
}
\caption{Impact of the \textbf{FaLB} entity alias augmentation step on REBEL and SynthIE. Inference is done on the full test set. The model is \preranker.
% Using entity augmentation is crucial, especially considering that real world use cases will likely heavily use aliases.
}
\label{table:alias-augment}
\end{table}

Expectedly, we observe that training with entity aliases allows us to link such OIE entity mentions more successfully than training without them. This was reflected on models trained on both REBEL and SynthIE. We further observe that this step hurts the linking of specific OIE entity mentions only moderately, suggesting that OIE linking methods could be trained to be robust w.r.t.~entity synsets. Finally, across all OIE slots and fact linking, the macro scores are significantly in favor of the model trained with entity aliases, on both datasets.

\begin{figure}[t]
\centering
\resizebox{1.0\columnwidth}{!}{
\includegraphics[width=1.0\columnwidth]{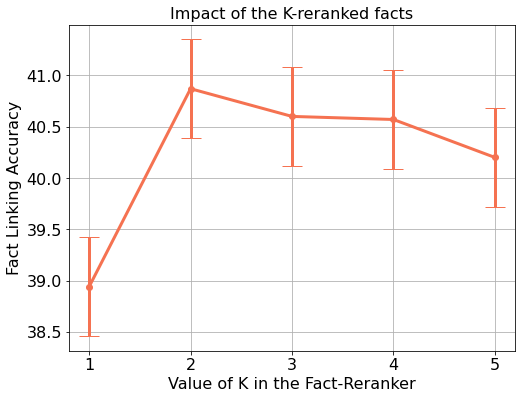}
}
\caption{Ablating the impact of K in the \reranker.}
\label{fig:reranker-k}
\end{figure}

% In addition, these models retain their linking performance on specific OIE entity mentions to a large extent.
% \subsection{Full Zero-Shot Setting}\label{sec:zero-shot}
% \begin{table}[t]
% \centering
% \resizebox{1.0\columnwidth}{!}{
% \begin{tabular}{lcccc} \toprule
%     \rowcolor{rowrow} {Method} & {Subject} & {Relation} & {Object} & {Fact} \\ \midrule
%     Slots pre-ranker (GN) & 43.89 \error{1.24} & 0.37 \error{0.15} & 45.64 \error{1.24} & 0.12 \error{0.09} \\
%     Slots pre-ranker (GN) + Rel augment & 46.38 \error{1.25} & 4.55 \error{0.52} & 49.38 \error{1.25} & 1.62 \error{0.32} \\ \bottomrule
% \end{tabular}
% }
% \caption{OIE slot-linking on fully zero-shot facts (i.e., both entities \& relations are unseen during training). Models trained on REBEL.}
% \label{table:zero-shot}
% \end{table}
\paragraph{Ablation Study: Importance of Fact-reranking.} We ablate the number of KG facts we rerank ($k$) with \reranker and report results in Fig.~\ref{fig:reranker-k}. We observe the highest fact linking accuracy when performing reranking using the top-2 highest scoring KG entries for each OIE slot, although performance is within 1 standard deviation for $k = 2,3,4$. If reranking $k=2$ facts, effectively, the \reranker constructs $2^3$ OIE-KG fact pairs, reranks the list, and returns the highest scoring KG fact.

\subsection{Detecting Out-of-Knowledge Graph OIEs}\label{sec:out-of-kg-detection}

\begin{table*}[t]
\centering
\resizebox{0.9\textwidth}{!}{
\begin{tabular}{rccccccccc} \toprule
    % \rowcolor{rowrow} {} & {} & \multicolumn{4}{c}{Benchmark Index} & \multicolumn{4}{c}{Entire Wikidata Index} \\
    \rowcolor{rowrow} {Method} & {Subject} & {Relation} & {Object} & {Fact} & {Subject} & {Relation} & {Object} & {Fact} \\ \midrule
    \rowcolor{row} {} & \multicolumn{4}{c}{\textit{Benchmark-Restricted Knowledge Graph}} & \multicolumn{4}{c}{\textit{Large Knowledge Graph}} \\ \midrule
    \multicolumn{9}{l}{\textit{\textbf{Testing data \underline{with} entity alias augmentation}}} \\
    Random & 50.0 \error{1.3} & \textbf{50.0 \error{1.3}} & 50.0 \error{1.3} & 12.5 \error{0.8} & 50.0 \error{1.3} & \textbf{50.0 \error{1.3}} & 50.0 \error{1.3} & 12.5 \error{0.8} \\
    % pRe-ranker (GN) w/ Entropy & Out-of-Knowledge Graph & 65.80 \error{0.48} & \NA & 55.52 \error{0.49} & \NA \\
    Confidence@1-based Heuristic & 63.8 \error{1.2} & \textbf{49.7 \error{1.2}} & 65.0 \error{1.2} & 21.5 \error{0.4} & 62.0 \error{1.2} & \textbf{49.7 \error{1.2}} & 62.7 \error{1.2} & 18.5 \error{0.4}\\
    Entropy-based Heuristic & \textbf{67.7 \error{1.2}} & \textbf{49.0 \error{1.2}} & \textbf{68.6 \error{1.1}} & \textbf{22.8 \error{0.4}} & \textbf{63.1 \error{1.1}} & \textbf{49.0 \error{1.2}} & \textbf{64.5 \error{1.1}} & \textbf{23.0 \error{0.4}} \\
    % pRe-ranker (GN) w/ Entropy & In-Knowledge Graph & 71.94 \error{0.44} & \NA & 67.11 \error{0.46} & \NA \\
    Query-Key-Value Cross-Attention & 62.4 \error{1.2} & \textbf{48.8 \error{1.0}} & 63.8 \error{1.2} & 17.2 \error{0.2} & 55.3 \error{1.2} & \textbf{49.2 \error{1.2}} & 56.8 \error{1.2} & 16.6 \error{0.2} \\ \midrule
    \multicolumn{9}{l}{\textbf{\textit{Testing data \underline{without} entity alias augmentation}}} \\
    Random & 50.0 \error{2.8} & \textbf{50.0 \error{2.8}} & 50.0 \error{2.8} & 12.5 \error{1.8} & 50.0 \error{1.3} & \textbf{50.0 \error{1.3}} & 50.0 \error{1.3} & 12.5 \error{1.8} \\
    Confidence@1-based Heuristic & 71.2 \error{2.3} & \textbf{49.6 \error{2.6}} & 72.1 \error{2.3} & \textbf{30.2 \error{1.5}} & 71.2 \error{2.5} & \textbf{49.6 \error{2.7}} & 69.1 \error{2.5} & 24.5 \error{1.5} \\
    Entropy-based Heuristic & \textbf{77.5 \error{2.3}} & \textbf{49.0 \error{2.7}} & \textbf{78.4 \small{$\pm$}} 2.2 & \textbf{29.9 \error{1.7}} & \textbf{73.3 \error{2.3}} & \textbf{49.1 \error{2.7}} & \textbf{72.8 \error{2.4}} & \textbf{27.0 \error{1.7}} \\
    % pRe-ranker (GN) w/ Entropy & In-Knowledge Graph & 71.94 \error{0.44} & \NA & 67.11 \error{0.46} & \NA \\
    Query-Key-Value Cross-Attention & 70.7 \error{2.5} & \textbf{49.6 \error{2.7}} & 70.9 \error{2.5} & 25.9 \error{1.0} & 59.8 \error{2.7} & \textbf{49.4 \error{2.7}} & 60.8 \error{2.5} & 19.7 \error{0.9} \\
    \bottomrule
    % Slots pre-ranker (GN) & \NA & Slot Linking & 43.89 \error{1.24} & 0.37 \error{0.15} & 45.64 \error{1.24} & 0.12 \error{0.09} \\ \bottomrule
\end{tabular}
}
\caption{Detection accuracy of Out-of-Knowledge Graph entities and predicates on the Out-of-KG split -- built on top of SynthIE. All models trained on REBEL with entity alias augmentation.
% All methods are very bad at detecting out-of-KG relations. They are better at detecting out-of-KG entities with the entropy-based heuristic performing best overall.
}
\label{table:out-of-kg}
\end{table*}
%\error{1.2} & 49.3 \error{1.3} & 65.0 $\small{\p

% 63.1 \error{1.1} & 51.8 $\error{1.2} & 64.5 \error{1.1} & 23.4 \error{0.4}

\paragraph{Setup.} Prior works which study the OIE to KG linking problem \cite{zhang2019openki, wood2021integrating, jiang2021cori} make the assumption that \emph{all} OIE slots that need to be linked are present in the KG, which is rarely the case in practice. Here, we study (i) whether OIE slot linking methods can be converted to Out-of-KG detectors; (ii) the performance impact of the KG size on the Out-of-KG detection task; and (iii) to what extent is it more difficult to recognize the presence or absence of aliased OIE entity mentions.

\paragraph{Datasets.} We create the Out-of-KG test split by imposing constraints such that we mimic an out-of-distribution setting. The constraints are: (i) both the KG entities and predicates referred by OIEs are unseen during training (i.e., not in the KG); (ii) the testing OIE-to-KG pairs do not come from the training data distribution: we use models trained on REBEL, but evaluate on SynthIE (which we use to create the Out-of-KG split). We measure Out-of-KG detection performance on the original OIEs,
% (as initially obtained by the OIE systems),
and OIEs with alias augmented entity slots.
% on out-of-KG split: (1) the split contains only the OIEs obtained from the sentences; (2) OIE entity slots are alias augmented.

\paragraph{Evaluation Protocol.} %We evaluate the extent to which the models can detect if the corresponding KG entity or relation an OIE slot should link to is present or not in the KG. 
We evaluate to what extent we can detect if an OIE slot refers to a concept outside of the KG. For each OIE slot, we add its corresponding concept (KG entity or predicate) to the KG, and score a hit if the model outputs a positive score for that slot. Conversely, for each OIE slot, we remove its corresponding entity or predicate from the KG, and score a hit if the model outputs a negative score. We report the averaged accuracy over the two scenarios for each OIE slot.

\paragraph{Methods.} We create three methods on top of \preranker trained on REBEL:
% with negatives sampled from the entire KG ($+ \text{GN}$):
\begin{inparaenum}[(i)]
\item \textsc{Confidence@1-based heuristic}: We get the cosine similarities of the top-5 predictions from \preranker, compute the softmax, and threshold the top-1 confidence;
\item \textsc{Entropy-based heuristic:} We also obtain the top-5 cosine similarities, however, after softmax normalization compute their entropy. Finally, we threshold the entropy to obtain the prediction;\footnote{For each method, we find the optimal threshold on a hold-out set which we build on top of the REBEL validation set.}
\item \textsc{Query-Key-Value Cross-Attention}: We train a lightweight query-key-value cross-attention module
% \footnote{The weights are initialized with the identity matrix -- at the start of training the \preranker embeddings are used as is. See Appendix~\ref{app:out-of-kg-models} for details.}
on top of the frozen \preranker embeddings. Given a query OIE slot embedding, the model attends over the KG entry embeddings representing the keys and values, and outputs a probability indicating the presence of the OIE slot in the KG
%We train the model to output a positive score if the corresponding KG entry is in the sampled subset, and negative otherwise. We obtain negatives by dropping KG entries (with 50\% probability), and randomly sample additional ones from the whole KG 
(See Appendix~\ref{app:out-of-kg-models} for details).
% During training, for each OIE slot, we drop the KG entry counterpart with 50\% probability and sample negatives from the whole KG. If the corresponding entry is in the sampled KG subset, we train the model to predict a positive score averaged over the KG graph subset, and negative otherwise.
\end{inparaenum}

\paragraph{Results.} We report the Out-of-KG detection performance in Table~\ref{table:out-of-kg}. First, we observe that none of the models we evaluate are able to recognize whether an OIE relation has a corresponding KG predicate, indicating that models do not cope with zero-shot relations. Intuitively, the number of entities is orders of magnitude more than the relations, thus the models learn features which generalize to unseen data. On the other hand, the number of relations is limited ($\sim$600 during training) and therefore, the models overfit on this limited set. Second, similar to our observations on the OIE linking task, detection of Out-of-KG slots is significantly more difficult on data which features entity aliases, and even more difficult on larger KGs. Lastly, the best performing model is an entropy-based heuristic on top of the \preranker output scores. Overall, our results indicate that Out-of-KG detection remains an open research problem.
\section{Related Work}\label{sec:rel-work}

\begin{table*}[t]
\centering
\resizebox{1.0\textwidth}{!}{
\begin{tabular}{rcccccccrr} \toprule
    \rowcolor{rowrow} {Dataset} & {Golden} &  {Manually} & {Multifaceted} & {Out-of-KG} & {Inductive} & {Transductive} & {Polysemous} & \# OIE Entities & \# OIE Relations   \\ 
    \rowcolor{rowrow} { } & {entities} & {validated rels.} & { } & { } & { } & { } &  { } & { } &  { } \\ \midrule
    FaLB (REBEL)  & \textcolor{forestgreen2}\cmark & \textcolor{forestgreen2}\cmark & \textcolor{forestgreen2}\cmark & \textcolor{forestgreen2}\cmark & \textcolor{forestgreen2}\cmark & \textcolor{forestgreen2}\cmark  & \textcolor{forestgreen2}\cmark & 936,655 & 159,597  \\
    FaLB (SynthIE)  & \textcolor{forestgreen2}\cmark & \textcolor{forestgreen2}\cmark & \textcolor{forestgreen2}\cmark & \textcolor{forestgreen2}\cmark & \textcolor{forestgreen2}\cmark & \textcolor{forestgreen2}\cmark & \textcolor{forestgreen2}\cmark  &  1,049,922  & 147,056 \\
    ReVerb45k & \textcolor{red}\xmark & \textcolor{red}\xmark & \textcolor{red}\xmark & \textcolor{red}\xmark & \NA & \NA & \textcolor{forestgreen2}\cmark & 28,798 & 21,925 \\
    %T-REx & \textcolor{red}\xmark & \textcolor{red}\xmark & \textcolor{red}\xmark & \textcolor{red}\xmark & \textcolor{red}\xmark & 0 & 0 & \textcolor{forestgreen2}\cmark & \textcolor{red}\xmark \\
    \bottomrule
\end{tabular}
}
\caption{Comparison of OIE-to-KG datasets: FaLB v.s.~ReVerb45k. As transductivity and inductivity are defined only w.r.t. what the model has observed during training, we leave these entries blank, because ReVerb45k has only validation and testing dataset.}
\label{table:datasets}
\end{table*}

Prior work \cite{zhang2019openki,wood2021integrating,jiang2021cori}---based on the ReVerb45k dataset \cite{Vashishth2018CESICO}---considers fact linking as inductive and polysemous for the entities, but perform inductive inference for the relations. They also assume the OIE entities are linked to the KG entities a priori. In contrast, FaLB requires linking of all OIE slots to KG entries, with multiple evaluation facets (including the out-of-KG setup).
% Gorjan: mislam deka e ok sega, fino ja sredi
% Further, they provide a single-facet of OIE evaluation, in which different error types are conflated.
% In contrast, our benchmark decouples these facets and allows evaluation of different OIE-to-KG linking scenarios: transductive, inductive, polysemous and out-of-KG.
In ReVerb45k \cite{Vashishth2018CESICO} the links from the OIE entities to the KG entities are not \textit{golden} (i.e., human labelled), but rather automatically obtained with an outdated entity linker \cite{lin2012entity}; thus, the poor performance of the entity linker caps the performance of the fact linking models. In addition, the number of benchmark KG predicates is unrealistically small (only 250) compared to modern KGs (e.g., Wikidata). In this work, we mitigate these issues and build a benchmark (i) with golden links to KG entities that also (ii) reflects the size of modern KGs. 

Furthermore, all prior work to date has relied on the strict assumption that \emph{all} OIE surface form slots have a corresponding reference KG entity or predicate \cite{wu2019scalable, jiang2021cori, zhang2019openki, josifoski2021genie}. This is obviously a false assumption, as any text corpus ``in the wild'' contains entities and predicates not present in even the largest of KGs \cite{Gashteovski2020OnAO}. See App.~\ref{app:rel-work} for detailed related work discussion.

Finally, most existing publicly available datasets do not address the problem of OIE-to-KG linking. Popular datasets like T-REx \cite{elsahar2018t} and REBEL \cite{cabot2021rebel}--if considered without modification--address only Text-to-KG alignment, thus lack the OIE component. Therefore, these datasets are not directly comparable to FaLB. To the best of our knowledge, the only publicly-available OIE-to-KG dataset is ReVerb45k \cite{Vashishth2018CESICO}, which, as indicated above, has several drawbacks. For detailed comparisons with and FaLB, see Table\ref{table:datasets}.
%making them unrealistic w.r.t.~the size of the current large KGs. For detailed comparison with and FaLB, see Tab.~\ref{table:datasets}.
%never the case with OIEs in practice. 
%rendering these evaluations unrealistic. 
%%%
%they assume that the entities of the surface facts are linked apriori, which is not a realistic assumption within the OIE paradigm. 
%, however, expressed in OIE - KG fact pairs containing entities and relation observed in the training data. That is, entities and relations constituting current knowledge graph facts.
%example, the entity mentions \emph{"Michael Jordan"} and \emph{"Chicago Bulls"} from  (Fig.~) are not seen at all during training in this scenario. Therefore, in this setting, we measure the extent to which OIE linking methods generalize beyond the training data.
%Another line of research that adopts inductive evaluation setup are works that perform link prediction (i.e., predict missing KG facts \cite{daza2021inductive, wang2021kepler,peng2022smile}) directly from text descriptions of KG nodes; which is, however, a fundamentally different task from fact linking. %, which we target in this work.   
%they do not consider the OIE component. Moreover, even though they deal with unseen entities and relations during training, their assumption is that the information is already extracted and canonicalized, which is rarely the case in realistic scenarios.

\section{Conclusion}
We shed light on the OIE to KG linking problem, allowing us \textit{to fuse the surface-form and opened-ended knowledge found in OIEs, with the canonical real-world KG facts}. We introduced a novel multifaceted benchmark which fixes prior work deficiencies, and proposed a set of task-specific baselines. Our experiments uncover that (i) linking inductive or polysemous OIEs to large KGs is challenging; (ii) we can learn OIE linking methods using only synthetic data; and (iii) detecting whether OIEs are Out-of-KG is an open research problem.

\section*{Acknowledgements}
We thank Dina Trajkovska for the help with the figures, and Mike Zhang for the fruitful discussions and feedback at the initial stages of the project.

\newpage
\section*{Limitations}
Notably, the set of models we explore ignore the KG structure to obtain KG entry embeddings. Leveraging the underlying graph structure should, in theory, yield representations which generalize better to zero-shot samples (e.g., as is the case with detecting out-of-KG relations). Such KG entry embeddings could be even trained offline (i.e., as a separate step) with standard KG embedding methods \cite{bordes2013translating}. 

Last but not least, all data, resources, and models used in this work are specific to the English language. Notice however, our approach can be readily extended to languages other than English, while Wikipedia and Wikidata have versions in other languages -- which we leave for future work.
% Additionally, our proposed approach might fail to work if the KG and the OIE facts are written in different scripts or if the textual representations of the KG are very different to the matching KG concepts.

\section*{Ethical Impact}
We are not aware of any direct ethical impact generated by our work. However, in general, care should be taken when applying our technology to sensitive use cases in high risk domains, such as healthcare.

\bibliography{anthology,custom}
\bibliographystyle{acl_natbib}

\clearpage

\appendix

\section{Detailed Discussion on Related Work}\label{app:rel-work}
In \S\ref{sec:fl}, we go over the problem statement, and in \S\ref{sec:rel-work} we discussed how the different facets of our benchmark relate to prior and closely related work. Here, we provide a broader discussion where we group the related work based on the problem they address, and provide a more detailed discussion w.r.t.~the differences with our work.

\paragraph{Open Information Extraction (OIE).} OIE methods extract structured surface-form factual information from natural language text data, in the form of \emph{(``subject''; ``relation''; ``object'')}-triples \cite{Banko2007OpenIE}.
%, gashteovski2017minie, bayat2022compactie, kotnis2022milie, ro2020multi, gashteovski-etal-2022-benchie}. We relate to these methods such that we propose a benchmark dataset, tasks, and models to link them to large-scale Knowledge Graphs \cite{vrandevcic2012wikidata}.
Such systems are typically either rule-based \cite{schmitz2012open,Corro2013ClausIECO,angeli2015leveraging,gashteovski2017minie,Lauscher2019MinScIECO,zopf2023fine} or neural-based \cite{stanovsky2018supervised,hohenecker2020systematic,kotnis2022milie,kotnis2022open,bayat2022compactie}. Recent research showed that neural OIE systems still fall behind rule-based systems \cite{Gashteovski2022BenchIEAF,Friedrich2022AnnIEAA}. To use the best of both worlds, in this work we use two rule-based systems: MinIE \cite{gashteovski2017minie} and Stanford OIE \cite{angeli2015leveraging}; and two neural systems: MILIE \cite{kotnis2022milie} and Multi$^2$OIE \cite{ro2020multi}.
% The task of Knowledge Graph population requires methods to populate an incomplete graph with the novel facts, otherwise not present in the graph. The facts, which need to be added in the graph, are of the form $[e_{subj}, r, e_{obj}]$, where $e_{subj}$ is the subject entity, $e_{obj}$ is the object entity, and $r$ is the relation.
% When adding a new fact to the knowledge graph, there are several possibilites:
% \begin{inparaenum}[(i)]
%     \item Both entities, and the relation already exists in the knowledge graph, but the fact does not. For such scenario, a ClosedIE method can invent a fact; 
%     \item Some of the entities, or the relation is not present in the knowledge graph.
% \end{inparaenum}
% We discuss the approaches currently present in the literature, discuss their limitation when applied for Knowledge Graph population. 

\paragraph{Closed Information Extraction.} Given an input text, ClosedIE methods extract a set of (subject; relation; object)-triples where each triple can be expressed within the predefined schema---fixed sets of entities and predicates---of the reference KG \cite{josifoski2021genie, distiawan2019neural, sui2021set, josifoski2023exploiting}. To ensure that each triple belongs to the KG, these methods prune candidates outside of the KG schema. Because of the pruning, these methods are unable to generate triples which contain entities or predicates that \textit{are not already in the KG}. In turn, the task we address---OIE linking to KGs---allows us to both link OIE to existing KG candidates, as well as detect \textit{novel} candidates which are outside of the KG schema.
% (in our work posed a task of detecting Out-of-KG entities and predicates).
% In turn, we design a specific task of detecting Out-of-KG entities and r
% , i.e., entities and relations are post-training the model.
% These methods can compose \textit{novel facts} strictly composed of entities and relations present in the knowledge graph. However, entities \& relations not present in the knowledge graph are simply ignored -- limiting their ability to populate the Knowledge Graphs with novel concepts.\par
% However, during inference, if the methods encounter \textit{novel entities and/or relations}, the they will not yield an extraction,

\paragraph{Open Information Extraction and Knowledge Graphs.} The context of OIE facts is used for many tasks for knowledge graphs, such as knowledge graph population \cite{Broscheit2017OpenIEFS,lin2020kbpearl,zopf2023fine}, (open) link prediction \cite{Broscheit2020CanWP,kotnis2023keyword}, entity linking \cite{nanni2019eal} and entity alignment \cite{friede2022context}. Therefore, the task of OIE fact linking to KGs is of great importance, as OIE facts can provide knowledge that is outside of the KG schema, as well as knowledge that aligns with the KG schema \cite{gashteovski2020aligning}. 
Several methods \cite{zhang2019openki,jiang2021cori, wood2021integrating} address the problem of linking OIEs to KGs. These methods, however, make the assumption that the subject and object OIE slots (which link to KG entities) are linked \textit{a priori}. These methods can link inductive OIE relations (outside of the training data), however, the assumption that the OIE entity slots are linked beforehand renders these methods not applicable for our task of linking (free-text) surface facts---also known as OIEs---to canonical large-scale KGs.
% make an unrealistic assumption that the OIE entities are aligned apriori with the knowledge graph entities. This is often not the case in practice, which limits their applicability.

\paragraph{Knowledge Graph Link Prediction.} These methods \cite{daza2021inductive, wang2021kepler, wang2022simkgc, peng2022smile} address the problem of Knowledge Graph population by predicting the missing facts in the KG given only the current set of facts. They assign higher scores to valid KG facts (composed of entities and predicates currently in the graph), which need to be added in the graph, and lower scores to erroneous facts (i.e., wrong fact proposals). Even though, during inference, these methods deal with entities and predicates which were unseen during training, an implausible assumption is made that information is already extracted and canonicalized as a KG facts, which is never the case in practice.\par

\paragraph{Multifaceted Evaluation.} NLP and KG tasks are typically evaluated on a held-out test set, by using evaluation frameworks that assign performance scores on a single value; e.g., accuracy \cite{Petroni2020KILTAB}. In recent years, researchers have observed that such evaluation protocols are somewhat limited \cite{Jain2023MultiDimensionalEO,Ye2021TowardsMF,Liu2021ExplainaBoardAE}, because they do not expose the particular types of problems that the models might have. Hence, with such evaluation protocols, the tested models are more difficult to diagnose when they make errors \cite{Ribeiro2020BeyondAB}. Following prior work on multifaceted OIE \cite{gashteovski-etal-2022-benchie} and multifaceted KG evaluation \cite{Meilicke2018FineGrainedEO,Rim2021BehavioralTO,Widjaja2022KGxBoardEA}, we propose FaLB: a multifaceted evaluation framework for fact linking. FaLB allows fine-grained evaluation that helps users to pinpoint the source of error (e.g., on which slot an examined model makes an error), which makes the benchmark more interpretable and useful for subsequent diagnostics of the models. In addition, our benchmark evaluates the performance of the models in different scenarios: inductive, transductive, polysemous, and out-of-KG. With this, the evaluation framework is more human-centric, in a sense that it can help users identify the model that they want for their needs instead of relying on single-score metrics \cite{Kotnis2022HumanCentricRF,Saralajew2022AHA}.

% To address these limitations, and accurately measure to what extent we can populate Knowledge Graphs with incoming and novel information about the real world, we propose the task of knowledge graph population using open information extractions. 
% construct a benchmark dataset, and propose XYZ tasks to measure the ability to ground Open Information Extractions -- facts represented in their surface, ambiguous form -- to Knowledge Graphs. 

\section{Benchmark Datasets Statistics}\label{app:benchmark-data-stats}
In Table~\ref{table:stats} we report statistics of the benchmark datasets we create on top of REBEL and SynthIE, using the Wikidata Knowledge Graph.

\begin{table*}[t]
\centering
\resizebox{1.0\textwidth}{!}{
\begin{tabular}{ccrrrr} \toprule
    \rowcolor{rowrow} {Backbone Dataset} & {Split Type} & {\# Total Samples} & {\# Unique Entities} & {\# Unique Predicates} & {\# Unique Facts} \\ \midrule
    REBEL & Training & 5,638,244 & 572,020 & 555 & 613,047 \\
    SynthIE & Training & 7,749,603 & 685,959 & 757 & 766,032 \\ \midrule
    REBEL & Full Validation & 421,547 & 48,103 & 303 & 41,092 \\
    SynthIE & Full Validation & 48,347 & 6,827 & 466 & 4,766 \\ \midrule
    % REBEL & Validation & 48103 & 303 & 41092 & 421547 \\ \midrule
    REBEL & Full Testing split & 427,961 & 48,807 & 314 & 41,767 \\
    SynthIE & Full Testing split & 240,300 & 28,660 & 635 & 22,661 \\ \midrule
    REBEL & Testing Transductivity & 241,995 & 14,730 & 192 & 13,496 \\
    REBEL & Testing Inductivity & 10,300 & 4,423 & 185 & 2,297 \\ %Inductive KG facts
    SynthIE & Testing Inductivity & 6,592 & 3,372 & 196 & 1,775 \\ %Inductive KG facts
    REBEL & Testing Polysemy & 15,339 & 3,504 & 117 & 3,290 \\
    SynthIE & Testing Out-of-KG & 1,604 & 443 & 106 & 264 \\ %Out-of-KG Detection
    \bottomrule
\end{tabular}
}
\caption{REBEL and SynthIE: overview of the number of samples in each dataset, number of unique KG entities, number of unique KG predicates, number of unique KG facts. Reported for all data splits of REBEL and SynthIE.}
\label{table:stats}
\end{table*}
\section{Implementation Details}\label{app:experimental}
We train all models for 10 epochs using AdamW with a learning rate of 5e-5 and weight decay of 1e-3. We use a RoBERTa \cite{liu2019roberta} model, distilled following the procedure of \citet{sanh2019distilbert}. The model consists of 6 layers, a hidden size of 768, and has 12 self-attention heads. To reduce the computational complexity and memory demands, we further linearly project the embeddings obtained from the RoBERTa model to a 200-dimensional latent space. When training the \preranker models, we initialize the temperature $\tau$ to 0.07 as per \citet{radford2021learning}. We train \preranker models with negatives that are sampled from the whole KG, where we sample 128 negative KG entities and 64 negative KG predicates. When training the \reranker, we sample negatives such that we corrupt the slots of the KG fact by replacing them with incorrect ones. Instead of choosing the negatives at random, for each KG entry, we find its top-10 most similar KG entries, and sample negatives from this subset. This ensures that the \reranker model learns how to refine the predictions of the \preranker.
We implement everything using PyTorch \cite{paszke2017automatic}, while we use HuggingFace transformers \cite{wolf2019huggingface} for the RoBERTa implementation. Lastly, we use Faiss \cite{johnson2019billion} to enable fast linking to Knowledge Graphs.

\section{Discussion on the Quality of REBEL and SynthIE}\label{app:rebel-vs-synthie}
While REBEL's entities are golden (i.e., obtained as the hyperlinks from Wikipedia abstracts which link to Wikipedia pages), the predicates between them are obtained using a set of heuristics. This leads to imbalanced data, where most predicates occur only few times, and others occur orders of magnitude more. Consequently, such issue is reflected in the OIE-to-KG fact linking data that we obtain. To address this problem of REBEL, \citet{josifoski2023exploiting} proposed SynthIE: a synthetically generated dataset which deals with the imbalance. 

\citet{josifoski2023exploiting} perform human evaluation to verify whether their synthetically generated dataset (of natural language sentences pair with KG facts) is of higher quality than REBEL. They obtain 44 data samples from REBEL, and generate a sentence using a Large Language Model (LLM) given only the KG triplets from the sample. Finally, they verify the triplet-set-to-text compatibility for both REBEL and SynthIE. Importantly, human evaluators find that the LLM generated sentences (i.e., the synthetic ones) are more compatible with the set of the KG facts (that is, the validity of the KG facts is higher in SynthIE compared to REBEL).

On the other hand, the issue we observe with SynthIE, is that due to the way the data is provided to LLM, the surface-form entities remain in their canonical text-form (i.e., their canonical denotation in the KG), and thus contain little variation. This diverges from the data that we find ``in the wild''. Critically, when people refer to entities in free-form natural language, they commonly use synonyms, aliases, abbreviations, nicknames, etc. For example, the former basketball player \emph{``Michael Jordan''} could be referred to as \emph{``Air Jordan''} and \emph{``M.J.''}.

To cope with this issue, we leverage the entity alias augmentation step in FaLB. By increasing the diversity of the OIE entity surface form, we are able to obtain a high quality OIE-to-KG fact linking synthetic dataset, thus the only human-annotated component remains to be the KG.

\subsection{Inductive Splits between Datasets}\label{app:inductive-splits}
Importantly, the inductive evaluation is testing OIE linking to KG facts that consist of entities which were not part of the models' training data. In \S\ref{sec:kg-population}\footnote{More precisely, in the paragraph titled \emph{``Results from Training on Synthetic Data''}}, we perform experiments by training models on REBEL and SynthIE, and then evaluate how well they perform on inductive data splits w.r.t.~each of the datasets. Namely, to obtain an inductive REBEL testing split w.r.t.~SynthIE, we find all testing samples from REBEL, which contain KG entities that are \textit{not part} of any SynthIE training samples. In turn, to obtain an inductive SynthIE testing split w.r.t.~REBEL, we find all SynthIE testing samples which contain KG entities that are \textit{not part} of any training samples from REBEL. We report statistics of the inductive splits w.r.t.~each dataset in Table~\ref{table:inductive-wrt-datasets}.

\begin{table}[t]
\centering
\resizebox{1.0\columnwidth}{!}{
\begin{tabular}{lcccc} \toprule
    \rowcolor{rowrow} {Dataset Type} & {\# Samples} & {\# Entities} & {\# Predicates} & {\# Facts} \\ \midrule
    REBEL w.r.t. Synthie & 26,937 & 8,781 & 181 & 5,304 \\
    SynthIE w.r.t. REBEL & 17,931 & 6,318 & 488 & 3,499 \\ \bottomrule
\end{tabular}
}
\caption{Overview of the total number of samples, number of unique KG entities, number of unique KG predicates, number of unique KG facts, and the. Reported for inductive splits w.r.t. each dataset.}
\label{table:inductive-wrt-datasets}
\end{table}

\section{Details on the Out-of-Knowledge Graph Detection Task}
In \S\ref{sec:out-of-kg-detection} we evaluate the ability of the models to detect whether an OIE slot (surface-form entity, or surface-form relation) is present in the Knowledge Graph. Intuitively, this task is more difficult than the OIE linking task, as the models need to generalize beyond the training data distribution to perform well on this task. Namely, when linking OIEs to a KG, all prior work \cite{zhang2019openki,jiang2021cori,wood2021integrating} makes the conjecture that the testing data (in the open-world) is independent and identically distributed (i.i.d.) w.r.t.~the training data, which is an invalid assumption in certain scenarios. Notably, the Out-of-KG OIEs lie outside of the training data distribution. Therefore, a model that performs well on the OIE linking task does not warrant high performance on the Out-of-KG detection task. We observed that this is the case (i.e., a model performs well on the linking task, but performs poorly on the Out-of-KG task) when detecting out-of-KG relations.

Therefore, to ensure that the Out-of-KG data is indeed out-of-distribution (as it would be the case in practice),
% To satisfy the aforementioned conditions when creating the Out-of-KG detection split,
we impose the following constraints: (i) We select OIE-to-KG pairs from SynthIE, while the ``backbone'' model we build on top of is trained on REBEL; (ii) All entities and predicates---which are part of the KG facts from the testing data---are not in the KG at the time of training the \preranker.

To perform the evaluation, for each OIE slot we either leave its corresponding KG entry outside of the KG, and score a hit if the models predict a \textit{negative score} for that slot; or, we perform imputation of the Out-of-KG entries (thus, they are now part of the KG), and score a hit if the models predict a \textit{positive score} for that slot. Finally, we report the averaged accuracy over the two scenarios for each OIE slot. Note that, in this case both micro- and macro-average yield the same number, because we have the same number of samples for each scenario.

\subsection{Out-of-Knowledge Graph Detection Models}\label{app:out-of-kg-models}
All models that we use for the Out-of-KG detection task in \S\ref{sec:out-of-kg-detection} are built on top of a \preranker, which is trained on REBEL. For all models, to obtain an out-of-KG indicator---True (1), or False (0)---we threshold the output of the models. For each model, we determine the optimal threshold (the confidence, or the entropy) on a hold-out validation dataset which we build on top of REBEL.
%\begin{itemize}
    \paragraph{\textsc{Confidence@1-based heuristic:}} We obtain the KG links for each of the OIE slots using the \preranker. The linking is characterized by the cosine similarity between the embeddings of each OIE slot and the KG entries. We then compute the softmax of the top-5 highest cosine similarities, and finally threshold the confidence@1 to obtain a prediction; such that a confidence $< T_c$ indicates an out-of-KG instance, and a confidence $> T_c$ indicates an instance inside the KG, where $T_c$ is the confidence threshold. We use $T_c = [0.235; 0.260; 0.235]$ for detecting out-of-KG subjects, relations and objects respectively.
    \paragraph{\textsc{Entropy-based heuristic:}} Similar to the confidence@1-based method, we obtain the cosine similarities with the \preranker. However, instead of using the top-1 probability, we obtain the entropy of the top-5 predictions. We finally threshold the entropy to obtain a prediction, such that an entropy $> T_e$ indicates out-of-KG instance, and an entropy $< T_e$ indicates an instance inside the KG, where $T_e$ is the entropy threshold. We use $T_e = [1.60; 1.58; 1.60]$ for detecting out-of-KG subjects, relations and objects respectively.
    \paragraph{\textsc{Query-Key-Value Cross-Attention:}} Using the \preranker embeddings, we train a lightweight query-key-value cross-attention module on top with weights that are initialized with the identity matrix---at the start of training the \preranker embeddings are used as is. Given a query OIE slot embedding, the model attends over the KG entry embeddings representing the keys and values, and outputs a sigmoid normalized score (i.e., a probability) indicating the presence confidence of the OIE slot in the KG. During training, to obtain in-batch negatives, for each OIE slot we drop the KG entry counterpart with 50\% probability. We obtain additional negatives by sampling KG entries from the whole KG, which do not match any of the OIE slots in the batch. We train the model using the binary cross-entropy loss, such that, if the corresponding KG entry for an OIE slot is in the sampled KG subset, the model predicts a positive score averaged over the KG graph subset, and negative otherwise. We use a uniform threshold of $T_a = 0.3$ for all three slots.
%\end{itemize}
\section{Data Quality}\label{app:data-quality}

%In what follows, we discuss the human evaluation study (\S\ref{sec:app-human-eval-steps}) as well as the detailed annotation guides (\S\ref{sec:app-annotation-guidelines}).

\begin{table*}[!ht]
\centering
\small

\begin{tabular}{lll} \toprule

    % \rowcolor{rowrow} {} & {} & \multicolumn{4}{c}{Benchmark Index} & \multicolumn{4}{c}{Entire Wikidata Index} \\
    \rowcolor{rowrow}  {KG fact (IDs)} & KG fact (names) & {OIE Triple} \\ \midrule
        \multicolumn{3}{l}{\textbf{Sentence:} \emph{``Hekimoğlu Ali Pasha Mosque was built between 1734–1735 in the Fatih district of Istanbul by}} \\ 
        \multicolumn{3}{l}{\hspace{1.55cm}\emph{\textcolor{green2}{Hekimoğlu Ali Pasha}, who \textcolor{red}{was born in} \textcolor{blue}{Istanbul} in 1689.''}} \\ 
        (\textcolor{green2}{Q1584693}; \textcolor{red}{P19}; \textcolor{blue}{Q406}) & (\textcolor{green2}{Hekimoğlu Ali Pasha}; \textcolor{red}{place of birth}; \textcolor{blue}{Istanbul}) & \emph{(\textcolor{green2}{``H. Ali Pasha''}, \textcolor{red}{``was born in''}, \textcolor{blue}{``Istanbul''})} \\
        \multicolumn{3}{l}{\textbf{Label:} correct} \\ \midrule
        %['Q1584693', 'P19', 'Q406']	['Hekimoğlu Ali Pasha', 'place of birth', 'Istanbul']	['Hekimoğlu Ali Pasha', 'was born in', 'Istanbul']
        \multicolumn{3}{l}{\textbf{Sentence:} \emph{``\textcolor{green2}{Pierre Hétu} (April 22, 1936 in \textcolor{blue}{Montreal} – December 3, 1998 in Montreal) was a conductor and pianist.''}} \\
        (\textcolor{green2}{Q3385492}, \textcolor{red}{P19}, \textcolor{blue}{Q340}) & (\textcolor{green2}{Pierre Hétu}; \textcolor{red}{place of birth}; \textcolor{blue}{Montreal}) & \emph{(\textcolor{green2}{``Pierre Hétu''}, \textcolor{red}{``April in''}, \textcolor{blue}{``Montreal''})} \\
        \multicolumn{3}{l}{\textbf{Label:} incorrect} \\
        
        %\multicolumn{3}{l}{\textbf{Sentence:} \emph{``\textcolor{green2}{Alis Guggenheim} (8 March 1896–2 September 1958) \textcolor{red}{was} a Swiss painter, and sculptor \textcolor{red}{born in} \textcolor{blue}{Lengnau}.''}} \\
        %(\textcolor{green2}{Q2647161}; \textcolor{red}{P19}; \textcolor{blue}{Q69429}) & (\textcolor{green2}{A. G.}; \textcolor{red}{place of birth}; \textcolor{blue}{Leng., Aargau}) & \emph{(\textcolor{green2}{``Alis G.''}; \textcolor{red}{``was sculptor born in''}; \textcolor{blue}{``Lengnau''})} \\
    \bottomrule
\end{tabular}

\caption{Example annotations for FaLB data. The annotator sees: (1) the input sentence; (2) the KG fact with the original IDs (thus, the user can further check the meaning of the entity and predicate); (3) the surface text of the KG fact; (4) the OIE triple, extracted from the input sentence. The first matching is labelled as \emph{``correct''}, because the KG fact semantically matches the OIE triple. The second matching is labelled as \emph{``incorrect''}, due to the incorrect OIE relation \emph{``April in''}.}
\label{table:data-samples}
\end{table*}

To assess FaLB's data quality, we performed manual human evaluation. In particular, we did the following steps: 
\begin{enumerate}
    \item We randomly selected 100 data points, where each data point contained information about the provenance sentence, a KG fact that is contained in the sentence, and a corresponding OIE surface fact that was extracted from the sentence.
    \item Two expert annotators annotated each data point independently of whether the KG fact matches the OIE extraction semantically (see two labelled examples in Table \ref{table:data-samples}). %(for details on the annotation guidelines, see \S\ref{sec:app-annotation-guidelines}). 
    \item We considered a data point as \emph{``correct''} only if the two annotators agreed that %(1) both the information in the KG fact and the OIE triple are indeed expressed within the provenance sentence $s$; (2) 
    the KG fact semantically matches the information in the OIE triple.
    \item We computed accuracy, inter-annotator agreement and Kohen's kappa score. 
\end{enumerate}

We found that 97\% of the data points are considered \emph{``correct''} by both annotators. We also observed that the inter-annotator agreement was high: the annotators agreed in 99\% of the cases, with high Kohen's kappa score \cite{mchugh2012interrater} of 0.80. Please refer to the supplementary material for the subset of samples which was provided to the expert annotators.

\end{document}